\documentclass[11pt]{article}

\usepackage[final]{acl}

\usepackage{times}
\usepackage{latexsym}

\usepackage[T1]{fontenc}

\usepackage[utf8]{inputenc}

\usepackage{microtype}

\usepackage{inconsolata}

\usepackage{graphicx}
\graphicspath{{figures/}}
\usepackage{amsmath}
\usepackage{algorithm}
\usepackage{algpseudocode}
\usepackage{float}

\title{Scaling Scientific Discovery Environments for Turn-Level Agentic RL}

\hypersetup{
  pdftitle={Scaling Scientific Discovery Environments for Turn-Level Agentic RL},
  pdfauthor={Yucheng Xu, Keyi Zhang, Yuyang Yu, Min Zhang, Shiyuan Meng, Pei Chu, Zhongying Tu}
}

\author{
  \textbf{Yucheng Xu}\textsuperscript{\ensuremath{\dagger}},
  \textbf{Keyi Zhang}\textsuperscript{\ensuremath{\dagger}},
  \textbf{Yuyang Yu},
  \textbf{Min Zhang},
  \\
  \textbf{Shiyuan Meng},
  \textbf{Pei Chu},
  \textbf{Zhongying Tu}\textsuperscript{*}
  \\
  Shanghai AI Lab
  \\
  \small{
    \textsuperscript{\ensuremath{\dagger}}Equal contribution
    \quad
    \textsuperscript{*}Corresponding author
  }
}

\begin{document}
\maketitle
\begin{abstract}
Large language model agents have shown promising capabilities in data-driven scientific discovery tasks, where an agent interacts with an execution environment and produces a statistical claim. Long-horizon scientific analysis remains constrained by the lack of process-supervised environments over real-world scientific data. This paper introduces \textbf{SciDisco}, a scalable framework for training \textbf{Sci}entific \textbf{Disco}very agents in process-verifiable environments. \textbf{SciTh\`eque} compiles hypotheses, datasets, hidden evidence graphs, and verifiers into task environments where analytical progress can be checked during interaction. DAG-grounded trajectory synthesis uses these environments to construct verifier-filtered multi-turn demonstrations. \textbf{DiscoPO} then uses the environment as the source of training signal, assigning turn-level credit to actions that produce verifiable analytical evidence. Experiments show that SciDisco-14B reaches state-of-the-art on hypothesis-driven scientific data analysis benchmarks.
\end{abstract}

\section{Introduction}
Data-driven scientific discovery is a hypothesis-driven mode of empirical research over existing scientific datasets. In human practice, it requires researchers to understand scientific context and dataset structure, select appropriate methods, execute statistical analyses, and draw verifiable conclusions. Existing LLMs and LLM-based agents remain unreliable in this setting. Prior work shows that models struggle with statistical method applicability and data-based causal reasoning~\citep{zhu2024statqa,liu2024qrdata}. Agentic code and reasoning scaffolds also achieve limited performance on data-driven discovery benchmarks~\citep{majumder2024discoverybench,gu2024blade,chen2025scienceagentbench}. These results suggest that data-driven scientific discovery should not be treated as one-shot question answering or single-turn code generation, but as an interactive process of executing, observing, revising, and verifying analyses over real data.

The core challenge is to train agents that learn robust interaction policies, rather than merely encode scientific knowledge or generate code. Prompts and hand-written scaffolds can describe the steps of an analysis workflow, but benchmark evidence suggests that prompting and generic multi-turn training do not reliably produce goal-directed interaction. In contrast, reward-trained tool use can outperform supervised tool traces in unfamiliar tool settings~\citep{abdulhai2023lmrlgym,qian2025toolrl,wang2025ragen}. Data-driven discovery agents must therefore learn from environmental feedback about when to inspect data, choose methods, test assumptions, revise analyses, and submit conclusions. This motivates training specialized discovery agents inside scientific data environments, rather than relying only on foundation models or inference-time workflow orchestration.

Such training requires environments that are both executable and verifiable. A discovery RL environment should provide data-driven goals, stateful code execution, observations, and automatic verification. Open scientific datasets are abundant, but they usually remain static files or metadata without task goals or verifiers. Existing scientific-discovery benchmarks provide valuable evaluation tasks and ground-truth criteria, but they do not expose hidden scientific evidence states for large-scale process-supervised training with persistent cross-turn state and verifier-grounded turn credit~\citep{majumder2024discoverybench,gu2024blade,chen2025scienceagentbench}. General agent and ML engineering environments support executable feedback and, in some cases, training, but their objectives target web interaction or ML engineering rather than real-data hypothesis verification~\citep{liu2023agentbench,chan2024mlebench,qiang2025mledojo,sandmle2026}.

Executable environments alone are still not enough for effective RL. Without a behavioral prior, early rollouts can be dominated by invalid tool calls, variable misreadings, and incomplete analyses. Data-driven scientific discovery also poses a long-horizon credit-assignment problem: final conclusions depend on schema understanding, method selection, code execution, diagnostics, and evidence interpretation, while outcome-only rewards cannot identify which turns produced valid analytical evidence~\citep{wang2025ragen,zeng2025turnlevelcredit,li2026tlgrpo,xie2026tips}.

These limitations point to a gap in agent training: data-driven scientific discovery needs an interface in which analytical progress can be executed, verified, and credited during interaction. \textbf{SciDisco} provides such a framework for data-driven scientific discovery agents. Its environment layer, \textbf{SciTh\`eque}, compiles open scientific datasets into verifiable hypothesis-driven environments. Hidden evidence DAGs support verified multi-turn demonstrations for SFT and provide process feedback that \textbf{DiscoPO} uses for turn-level credit assignment in agentic RL. SciDisco therefore aligns task construction, behavior imitation, and policy optimization within realistic scientific data environments.

The paper makes three contributions:
\begin{enumerate}
    \setlength{\itemsep}{0.15em}
    \setlength{\parsep}{0pt}
    \setlength{\parskip}{0pt}
    \item \textbf{SciTh\`eque} builds scalable discovery environments from open scientific datasets with hypotheses, sandboxed execution, and verifiers.
    \item DAG-grounded trajectory synthesis cold-starts agent behavior using verified multi-turn demonstrations.
    \item \textbf{DiscoPO} assigns turn-level credit to verified evidence-producing processes instead of only final answers.
\end{enumerate}

\section{Related Work}
\label{sec:related-work}

\subsection{Discovery Agents and Environments}
Prior work on data-science and scientific-discovery agents follows two broad paradigms. The first improves data-specific capabilities, including tabular understanding, database operations, statistical reasoning, and quantitative reasoning over data~\citep{zhu2024statqa,liu2024qrdata}. The second moves beyond single-turn QA by building agentic workflows with ReAct-style interaction, code interpreters, multi-agent scaffolds, or learned analysis trajectories. These systems often rely on strong closed-source models, leaving the underlying training environment difficult to reuse~\citep{guo2024dsagent,zhang2023datacopilot,hong2024datainterpreter,li2024autokaggle,zhang2025deepanalyze,qiao2025datamind}. This makes it difficult to separate model capability from the executable interaction and verification substrate used to produce scientific conclusions.

 Discovery benchmarks evaluate agents on real data, open-ended questions, code execution, and conclusion checking~\citep{majumder2024discoverybench,gu2024blade,chen2025scienceagentbench,dabstep2025}. Interactive science environments support exploration in text-based or virtual worlds~\citep{wang2022scienceworld,osullivan2024discoveryworld}. General agent and MLE environments cover tool use, web interaction, executable feedback, and training efficiency~\citep{liu2023agentbench,zhou2023webarena,chan2024mlebench,qiang2025mledojo,sandmle2026}. These settings are valuable, but they are not built around real-data hypothesis verification with persistent execution state and verifier-grounded process feedback. SciTh\`eque constructs executable, verifiable discovery environments that connect data, hypotheses, code execution, and reusable multi-turn feedback, so the same environment can define tasks, filter trajectories, and provide RL signals. This shifts the role of an environment from a static evaluation artifact to a reusable training interface for scientific agent learning.

\subsection{Reinforcement Learning for LLM Agents}
Beyond prompting and SFT, recent work trains LLM agents with reinforcement learning. GRPO-style RLVR uses group-relative advantages to avoid explicit critics and train models on verifiable reasoning or coding tasks~\citep{shao2024deepseekmath,guo2025deepseekr1}. Later variants improve sampling, clipping, normalization, length bias, and self-distillation routing~\citep{yu2025dapo,liu2025drgrpo,srpo2026}. These methods provide a strong optimization backbone for agent training, but their rewards usually remain at the response or trajectory level.

For multi-turn agents, recent work assigns credit at step, turn, or segment granularity, or calibrates intermediate decisions with hindsight, reflection, or trajectory correction~\citep{feng2025gigpo,zeng2025turnlevelcredit,li2026tlgrpo,zong2026at2po,guo2025spo,lu2026hisr,wang2025steca,wang2025stepsearch}. These methods recognize that final rewards cannot identify which intermediate actions advanced the task. However, their signals often depend on repeated states, tree search, segmenters, hindsight models, reflection, or task-specific step rewards, rather than evidence produced by executed scientific analyses. DiscoPO instead uses scientific verifiers to credit turns that produce validated analytical evidence, aligning turn-level RL with hypothesis verification.

\section{Problem Formulation}
\label{sec:problem-formulation}
A scientific discovery task is formulated as a hypothesis-driven data analysis problem. Given a dataset, the task is to evaluate a scientific hypothesis, produce statistical findings, and support those findings with executable analysis.

Let $\mathcal{U}=\{u_j\}_{j=1}^{N}$ denote a collection of such tasks. Each task is represented as:
\begin{equation}
u_j=(h_j,d_j,v_j),
\label{eq:task-formulation}
\end{equation}
where $h_j$ specifies the hypothesis or analysis goal, $d_j$ is the dataset associated with the task, and $v_j$ denotes the task verifier.

Each task $u_j$ is paired with an environment $e_j$, and the corresponding environment collection is $\mathcal{E}=\{e_j\}_{j=1}^{N}$. The environment exposes $h_j$ and $d_j$, executes the agent's actions in a sandbox, and returns observations, while task-internal verification state may remain hidden. The interaction between policy $\pi_\theta$ and environment $e_j$ can therefore be modeled as a partially observable Markov decision process (POMDP), producing a trajectory $\tau_j$:
\begin{equation}
\tau_j=(o_{j,0},a_{j,1},o_{j,1},\ldots,a_{j,T_j},o_{j,T_j}),
\label{eq:trajectory-formulation}
\end{equation}
where $o_{j,0}$ contains the initial task observation, actions $a_{j,t}$ can be executable analysis or terminal submission, and observations $o_{j,t}$ contain environment feedback.

A final reward provides only sparse credit assignment for a scientific discovery trajectory. It cannot distinguish an evidence-building trajectory from a one-shot shortcut that happens to produce the same submitted findings.

Let \(r_{j,t}\) denote the process reward assigned to turn \(t\), measuring whether that turn contributes verifiable progress toward the task findings. The training goal is to maximize the accumulated process reward:
\begin{equation}
R_j(\tau_j)=\sum_{t=1}^{T_j} r_{j,t},
\label{eq:process-return}
\end{equation}
and the optimization objective is:
\begin{equation}
J(\theta)=\mathrm{E}\left[R_j(\tau_j)\right],
\label{eq:policy-objective}
\end{equation}
where the expectation is over tasks sampled from $\mathcal{U}$ and trajectories generated by $\pi_\theta$ in their paired environments.

\section{Methodology}
\label{sec:methodology}
\textbf{SciDisco} turns the process-reward formulation in Section~\ref{sec:problem-formulation} into a three-stage post-training pipeline for scientific discovery agents. \textbf{SciThèque} compiles scientific datasets into sandboxed task environments with task-local hidden evidence DAGs. DAG-grounded trajectory synthesis uses these environments to produce verifiable multi-turn demonstrations for SFT cold-start. \textbf{DiscoPO} then optimizes the policy with turn-level credit from environment-verified scientific progress. The task-local hidden evidence DAG is the shared interface across these stages: it defines what can be synthesized, what counts as an accepted transition during online interaction, and how progress is credited during RL.

\begin{figure*}[t]
\centering
\includegraphics[width=\linewidth]{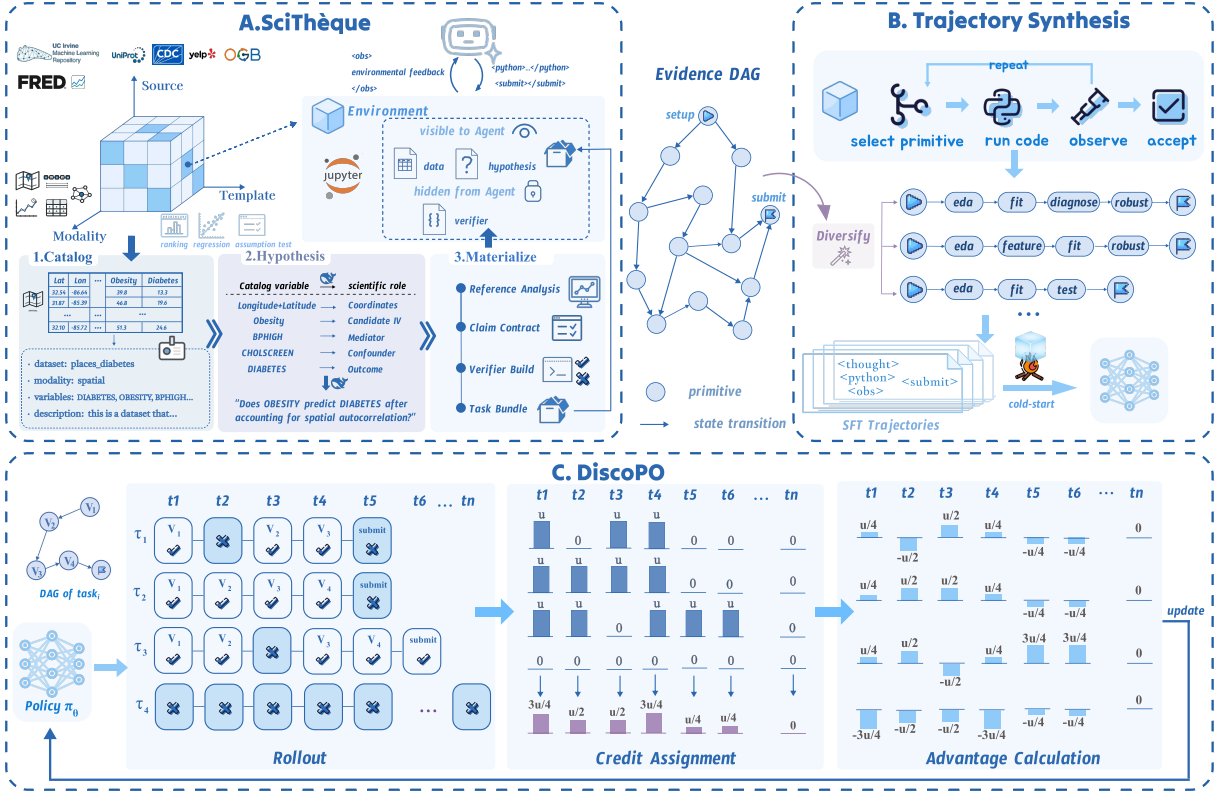}
\caption{The pipeline of \textbf{SciDisco}: 1) \textbf{SciTh\`eque} builds a scalable collection of data--hypothesis--verifier environments from scientific datasets; 2) \textbf{Trajectory Synthesis} traverses the evidence DAG through executable analysis primitives and keeps accepted state transitions as multi-turn SFT demonstrations; 3) \textbf{DiscoPO} defines turn-level rewards by increases in verified DAG progress, assigning credit to evidence-producing turns.}
\label{fig:scidisco-overview}
\end{figure*}

\subsection{SciThèque}
\label{sec:scitheque}
SciThèque instantiates the environment collection $\mathcal{E}$ defined in Section~\ref{sec:problem-formulation}. It constructs tasks $u_j=(h_j,d_j,v_j)$ from source datasets and hypothesis templates, then mounts each task in a sandbox runtime to form an environment $e_j$. The agent sees $h_j$, data files $d_j$, and observations $o_{j,t}$, while the verifier logic, evidence DAG, and progress state remain hidden. Appendix~\ref{app:env-details} reports the data-source distribution and materialized environment contract.

\noindent\textbf{Catalog construction.} SciThèque first builds candidate dataset entries $c_i=(s_i,d_i,m_i,\rho_i)$, where $s_i$ is the source, $d_i$ the dataset, $m_i$ the modality, and $\rho_i$ a structural profile recording provenance, licence metadata, file structure, size, and analysis-relevant fields. Inexpensive gates remove unsuitable datasets before task creation, and each surviving dataset is classified as tabular, temporal, graph, spatial, or sequence data. Appendix~\ref{app:env-details} reports source citations and snapshot counts.

\noindent\textbf{Template-guided hypothesis generation.} For each modality $m$, SciThèque maintains a template library $\mathcal{T}_m$ indexed by statistical pattern and outcome type. Each catalog entry is matched to templates allowed by $\rho_i$, and an external LLM proposes constrained specifications $\mathcal{H}_i=\{(T_{i,k},\eta_{i,k})\}_{k=1}^{K_i}$, where $T_{i,k}\in\mathcal{T}_{m_i}$ is an applicable template and $\eta_{i,k}$ instantiates it on $d_i$. The LLM proposes candidates but does not certify validity; retained specifications from all $\mathcal{H}_i$ form the materialization pool.

\noindent\textbf{Environment materialization.} Materialization turns each retained specification into a training environment. SciThèque runs the reference analysis on $d_i$ under the selected template and accepts the specification only if the analysis executes and yields verifier-backed findings. The accepted specification becomes $u_j=(h_j,d_j,v_j)$, then is paired with sandbox runtime $\mathcal{S}_j$ to form $e_j=(u_j,\mathcal{S}_j)$.

The verifier $v_j$ is grounded in a hidden evidence DAG:
\begin{equation}
g_j=(\mathcal{V}_j,\mathcal{A}_j),
\label{eq:scitheque-dag}
\end{equation}
Each node in $\mathcal{V}_j$ defines a verifiable scientific state transition and is typed by a process primitive, such as data inspection, model fitting, diagnostic checking, or robustness analysis. The primitive type determines the admissible family of analysis operations for that node, so a turn is verified against the scientific role of the transition rather than arbitrary code execution. The directed edges $\mathcal{A}_j$ encode prerequisite relations, so only frontier nodes are eligible at a given turn. The DAG is hidden from the agent and defines the progress state used by $v_j$ to verify intermediate work. Appendix~\ref{app:dag-acceptor-details} details the visibility split, evidence-graph record, progress rule, and leakage controls.

To assess whether the verifier design matches domain-level scientific practice, 10 domain experts covering the scientific domains represented in the environment corpus manually reviewed the evidence primitives, prerequisite structure, and acceptance criteria used by the hidden evidence graphs. Appendix~\ref{app:dag-acceptor-details} reports the review protocol and audit scope.

The materialized output is therefore an environment specification, not a static question-answer pair. Each task stores its data files, instructions, verifier, task-local evidence DAG, final checker, and reproducibility metadata.

\subsection{DAG-Grounded Trajectory Synthesis}
\label{sec:dag-grounded-trajectory-synthesis}
DAG-grounded trajectory synthesis uses SciThèque environments to construct SFT demonstrations for model cold-start. Although the evidence DAG defines verifiable scientific state transitions, a base policy may not know how to realize them as executable analysis turns. SFT therefore places the policy in the right interaction regime before RL begins. Appendix~\ref{app:sft-details} reports the synthesis stages, filtering criteria, and retained corpus snapshot.

For each materialized task $u_j$ with hidden evidence DAG $g_j$, the synthesizer constructs a private transition schedule $\sigma_{j,\ell}$ over the required DAG nodes. This schedule specifies the verified transition executed at each step while respecting the prerequisite structure of $g_j$. For each scheduled node, the synthesizer selects an executable program variant consistent with the node's process primitive. The schedule and program choices remain hidden. The transcript exposes only assistant reasoning \texttt{<thought>}, code actions \texttt{<python>}, sandbox observations \texttt{<obs>}, and a final \texttt{<submit>} action.

Each synthesized trajectory uses a fresh sandbox runtime, where the scheduled \texttt{<python>} actions are executed. After each execution, the verifier $v_j$ checks whether the visible observation establishes the scheduled transition in $g_j$. A trajectory is rejected if the code fails, produces no verifiable evidence, matches the wrong frontier node, violates the primitive process constraint, or accesses hidden verifier files. Each retained SFT turn therefore corresponds to an accepted scientific state transition rather than an arbitrary code fragment.

After all scheduled transitions are accepted, synthesis closes the transcript with a \texttt{<submit>} action. The submission records findings already established through verified transitions rather than introducing a separate terminal reward. Let $\mathcal{T}_{\mathrm{syn}}$ be the pool of candidate trajectories produced by synthesis, and let $\tau_{j,\ell}$ denote the $\ell$-th candidate trajectory for task $u_j$. A candidate is retained in $\mathcal{D}_{\mathrm{SFT}}$ only when the required nodes in $g_j$ have been completed and the submitted findings are consistent with the accumulated verifier state:
\begin{equation}
\mathcal{D}_{\mathrm{SFT}} = \{\tau_{j,\ell}\in\mathcal{T}_{\mathrm{syn}} \mid A_j(\tau_{j,\ell},g_j)=1\}.
\label{eq:sft-set}
\end{equation}
Here $A_j(\tau_{j,\ell},g_j)=1$ indicates that the trajectory completes the required DAG transitions in a valid order and submits the findings verified during those transitions.

For the same task, synthesis can retain several verifier-valid trajectories rather than a single canonical solution path. These trajectories may differ in transition schedule, executable program variant, finding-reporting order, and reasoning style, while sharing the same hidden DAG and verifier constraints. Training-data conversion removes scheduler labels, hidden node identifiers, reference values, verifier tolerances, and generator-only helper variables. This cold-start stage teaches the policy to reason through a scientific state-transition chain. DiscoPO can then optimize verified scientific progress from a policy that already follows the basic scientific behavior.

\subsection{DiscoPO}
\label{sec:discopo}

DiscoPO trains the policy in SciTh\`eque environments. At each turn within a trajectory, the agent emits either a \texttt{<python>} cell or a \texttt{<submit>} action. The sandbox presents each task as an agent workspace, and its persistent kernel lets later turns reuse earlier variables, models, and files. Each observation provides evidence for verifying the next state transition.

GRPO computes a group-normalized advantage from trajectory-level rewards and broadcasts that advantage to all generated tokens of the trajectory~\citep{shao2024deepseekmath}. DiscoPO keeps the same group-relative normalization but computes the advantage from verifier-gated progress at each interaction turn.

\begin{algorithm}
\small
\caption{DiscoPO}
\label{alg:discopo}
\begin{algorithmic}[1]
\Require task environments $\{e_j\}$, hidden DAGs $\{g_j\}$, old policy $\pi_{\mathrm{old}}$, group size $G$, clip parameters $\epsilon_{\mathrm{low}}$, $\epsilon_{\mathrm{high}}$
\For{each training iteration}
    \State sample a batch of task environments
    \For{each sampled environment $e_j$}
        \State sample $G$ rollouts $\{\tau_{j,k}\}_{k=1}^{G}$ with $\pi_{\mathrm{old}}$
        \For{each rollout $k=1,\ldots,G$}
            \State initialize $C_{j,k,0} \gets \emptyset$
            \For{each turn $t$ in $\tau_{j,k}$}
                \State use $v_j$ to update $C_{j,k,t+1}$ from $C_{j,k,t}$
                \State compute $r_{j,k,t}$ by Eq.~\ref{eq:discopo-turn-credit}
            \EndFor
        \EndFor
        \State compute $b_{j,t}$ and $A_{j,k,t}$ by Eqs.~\ref{eq:discopo-turn-baseline} and~\ref{eq:discopo-advantage}
        \State assign $A_{j,k,t}$ to generated tokens in turn $t$ and mask observations
    \EndFor
    \State update $\pi_\theta$ with the clipped surrogate over turn-level advantages
    \State set $\pi_{\mathrm{old}}\gets\pi_\theta$
\EndFor
\end{algorithmic}
\end{algorithm}

\noindent\textbf{Verifier-gated state transitions.} For task $u_j$, let $g_j=(\mathcal{V}_j,\mathcal{A}_j)$ be the hidden evidence DAG, and let $\mathcal{V}^{\mathrm{exec}}_j\subseteq\mathcal{V}_j$ denote the required DAG nodes verified through \texttt{<python>} turns. For rollout $k$ of task $j$, $C_{j,k,t}$ records the DAG nodes credited before turn $t$. Here $j$ follows Section~\ref{sec:scitheque} and indexes the materialized task, while $k$ indexes one trajectory in the rollout group used for group-relative advantage. Appendix~\ref{app:prompt-rollout-example} shows an example rollout trajectory.

At each turn, the verifier $v_j$ considers only frontier nodes, meaning uncredited nodes whose prerequisites have already been completed. The verifier then checks whether the current \texttt{<python>} turn matches the process primitive assigned to one frontier node. A turn is accepted only when the observation provides evidence for exactly one such node and the executed code satisfies its node-local process check.

Turns are rejected when they provide no observable evidence, match no frontier node, match multiple frontier nodes, repeat completed work, or collapse several analyses into one action. If a turn is accepted, the matched node is added to the hidden completed-node state; otherwise, the state is left unchanged. The \texttt{<submit>} action is checked against the verifier and can be accepted only after the required nodes have been credited. Appendix~\ref{app:failure-cases} gives concrete rejection categories and their training consequences.

\noindent\textbf{Credit assignment.} For task $u_j$, let $N_j=|\mathcal{V}^{\mathrm{exec}}_j|$ be the number of required executable primitive transitions, and let $s^{\mathrm{sub}}_j$ denote the terminal submit node. The set of progress nodes is $\mathcal{P}_j=\mathcal{V}^{\mathrm{exec}}_j\cup\{s^{\mathrm{sub}}_j\}$. DiscoPO defines the DAG-progress potential over credited-node sets as:
\begin{equation}
\Phi_j(C)=\frac{|C\cap\mathcal{P}_j|}{|\mathcal{P}_j|}.
\label{eq:discopo-potential}
\end{equation}
The per-turn process reward is the potential difference:
\begin{equation}
r_{j,k,t}=\Phi_j(C_{j,k,t+1})-\Phi_j(C_{j,k,t}).
\label{eq:discopo-turn-credit}
\end{equation}
When a \texttt{<python>} turn is accepted, it adds one previously uncredited primitive node. A valid \texttt{<submit>} action adds the terminal submit node once all required primitives have been completed. Turns that do not add a new progress node receive zero reward. Equivalently:
\begin{equation}
r_{j,k,t} =
\left\{
\begin{array}{ll}
\frac{1}{N_j+1}, & \mbox{accepted primitive/submit,}\\
0, & \mbox{otherwise.}
\end{array}
\right.
\label{eq:discopo-turn-reward}
\end{equation}

For each task environment, a group of $G$ rollouts $\{\tau_{j,k}\}_{k=1}^{G}$ is sampled from the old policy $\pi_{\mathrm{old}}$. Let $T_{j,k}$ be the length of rollout $k$. Rewards from terminated rollouts are zero-padded when computing the same-turn baseline:
\begin{equation}
\widetilde r_{j,k,t}=\left\{
\begin{array}{ll}
r_{j,k,t}, & t\leq T_{j,k},\\
0, & t>T_{j,k}.
\end{array}
\right.
\label{eq:discopo-zero-pad}
\end{equation}
This same-turn baseline compares whether rollouts for the same task make verifier progress at interaction index $t$; it does not require different rollouts to complete the same DAG node at that index. DiscoPO then applies turn group normalization. The same-turn baseline is:
\begin{equation}
b_{j,t} =
\frac{1}{G}\sum_{k=1}^{G}\widetilde r_{j,k,t}.
\label{eq:discopo-turn-baseline}
\end{equation}
The turn advantage is:
\begin{equation}
A_{j,k,t} =
\widetilde r_{j,k,t}-b_{j,t}.
\label{eq:discopo-advantage}
\end{equation}

\begin{table*}[!t]
\centering
\small
\setlength{\tabcolsep}{3.5pt}
\begin{tabular}{lc@{\hspace{10pt}}ccc@{\hspace{10pt}}cc}
\hline
\rule[-1.05ex]{0pt}{3.ex}\textbf{Model} & \multicolumn{1}{c}{\textbf{DiscoveryBench}} & \multicolumn{3}{c}{\textbf{DABStep}} & \multicolumn{2}{c}{\textbf{DataSciBench}} \\
\cline{2-2}\cline{3-5}\cline{6-7}
 \rule[-1.05ex]{0pt}{3.2ex} & Average HMS & Easy & Hard & Average & Success Rate & Completion Rate \\
\hline
\multicolumn{7}{c}{\rule[-1.05ex]{0pt}{3.2ex}\textit{Proprietary Models}} \\
\hline
GPT-4o & 13.2 & 66.7 & 6.1 & 15.8 & 66.3 & 68.4 \\
GPT-5 Mini & \textbf{26.9} & 68.1 & 15.3 & 23.8 & 45.9 & 63.4 \\
Claude-Sonnet-4 & 23.2 & 81.9 & 19.8 & \textbf{29.7} & \textbf{67.6} & \textbf{76.6} \\
\hline
\multicolumn{7}{c}{\rule[-1.05ex]{0pt}{3.2ex}\textit{Open-source Models}} \\
\hline
DeepSeek-V4-Flash & 28.3 & 79.2 & 12.2 & \textbf{22.9} & \textbf{60.8} & \textbf{72.1} \\
Intern-S1-Pro & \textbf{34.7} & 65.3 & 4.8 & 14.5 & 59.5 & 67.6 \\
Qwen3-8B & 11.3 & 47.2 & 3.2 & 10.2 & 31.5 & 48.4 \\
Qwen3-14B & 20.0 & 55.6 & 4.2 & 12.4 & 32.4 & 45.9 \\
Qwen3-32B & 24.1 & 59.7 & 4.5 & 13.3 & 36.0 & 52.4 \\
\hline
\multicolumn{7}{c}{\rule[-1.05ex]{0pt}{3.2ex}\textit{Related Methods}} \\
\hline
DeepAnalyze-8B & 25.9 & 70.8 & 32.8 & \textbf{38.9} & 46.8 & 57.3 \\
AutoSDT-Coder-14B & 7.3 & 56.9 & 2.7 & 11.3 & 1.0 & 1.3 \\
DataMind-14B & 25.5 & 68.1 & 3.2 & 13.6 & 20.3 & 24.8 \\
SciDisco-14B & \textbf{35.2} & 62.5 & 9.3 & 17.8 & \textbf{56.2} & \textbf{61.0} \\
\hline
\end{tabular}
\caption{Main results on DiscoveryBench, DABStep, and DataSciBench. Performance is reported as percentages. The best results for each model group are highlighted in bold.}
\label{tab:main-results}
\end{table*}

\noindent\textbf{Policy optimization.} The policy update assigns $A_{j,k,t}$ to policy-generated tokens in turn $t$, masks observation tokens, and applies a PPO clipped surrogate at turn granularity~\citep{schulman2017ppo}. Let $\mathcal{I}_{j,k,t}$ denote the generated-token positions in that turn, let $\rho_i(\theta)$ be the old-policy likelihood ratio for token $i$, and let $\bar{\rho}_i(\theta)$ be its clipped value. DiscoPO optimizes:
\begin{equation}
\begin{aligned}
\mathcal{L}_{\mathrm{DiscoPO}}(\theta)
&=
-\mathrm{E}_{j,k,t}
\left[
\frac{1}{|\mathcal{I}_{j,k,t}|}
\sum_{i\in\mathcal{I}_{j,k,t}} s_i(\theta)
\right],\\
s_i(\theta)
&=\min\left(
\rho_i(\theta)A_{j,k,t},
\bar{\rho}_i(\theta)A_{j,k,t}
\right).
\end{aligned}
\label{eq:discopo-objective}
\end{equation}
The optimizer is standard; the change is that verifier-gated DAG progress supplies turn-level token weights instead of a scalar trajectory reward. Groups whose turn advantages are identically zero are filtered before optimization because they provide no group-relative policy-gradient signal. Appendix~\ref{app:discopo-objective-details} expands the token-level notation.

\section{Experiments}
\label{sec:experiments}

The experiments evaluate whether SciDisco improves data-driven scientific discovery and executable scientific data analysis. The evaluation covers three benchmark regimes: hypothesis generation over scientific datasets, multi-step business data analysis, and programmatically checked data-science workflows.

\subsection{Benchmarks}
\label{sec:benchmarks}

\noindent\textbf{DiscoveryBench.} DiscoveryBench tests whether an agent can produce data-driven scientific hypotheses from a dataset and a discovery goal~\citep{majumder2024discoverybench}. Table~\ref{tab:main-results} reports the average Hypothesis Match Score (HMS), which aligns predicted and gold sub-hypotheses along context, variable, and relationship dimensions. GPT-5 Mini was used as the judge model during evaluation.

\noindent\textbf{DABStep.} DABStep contains 450 realistic multi-step data-analysis tasks over business datasets, split into 72 easy and 378 hard cases~\citep{dabstep2025}. Its tasks emphasise business data analysis, where an agent completes operational analyses over enterprise-like tables, reports, visualisations, or decision-support queries. Table~\ref{tab:main-results} reports easy accuracy, hard accuracy, and the task-count-weighted average.

\noindent\textbf{DataSciBench.} DataSciBench evaluates end-to-end data-science workflows with programmatic Task--Function--Code checks over data preparation, analysis, modelling, visualisation, and insight tasks~\citep{zhang2025datascibench}. Table~\ref{tab:main-results} reports the benchmark-level success rate and completion rate, while Table~\ref{tab:ablation-results} further uses the DataSciBench breakdown to report fine-grained F1--F5 scores for data preparation, plot validity, data exploration, data visualization, and data modeling.

\begin{table*}[!t]
\centering
\footnotesize
\setlength{\tabcolsep}{2pt}
\begin{tabular*}{\textwidth}{@{\extracolsep{\fill}}lcccccccc@{}}
\hline
\rule[-1.05ex]{0pt}{3.4ex}\textbf{Setting} & \multicolumn{2}{c}{\textbf{Coarse-grained Metrics}} & \multicolumn{6}{c}{\textbf{Fine-grained Metrics}} \\
\cline{2-3}\cline{4-9}
\rule[-1.05ex]{0pt}{3.4ex} & \textbf{\shortstack{Success\\Rate}} & \textbf{\shortstack{Completion\\Rate}} & \textbf{VLM} & \textbf{\shortstack{F1: Data\\Preparation}} & \textbf{\shortstack{F2: Plot\\Validity}} & \textbf{\shortstack{F3: Data\\Exploration}} & \textbf{\shortstack{F4: Data\\Visualization}} & \textbf{\shortstack{F5: Data\\Modeling}} \\
\hline
Qwen3-14B & 32.4 & 45.9 & / & 16.1 & 41.1 & 29.3 & 23.6 & 22.3 \\
+ SFT & 46.8 & 53.3 & / & 37.9 & 51.8 & 32.8 & 50.9 & 23.4 \\
+ SFT + GRPO & 49.5 & 55.4 & / & 39.8 & 53.6 & 34.5 & 52.7 & 26.3 \\
+ SFT + DiscoPO & 56.3 & 61.0 & / & 43.0 & 57.1 & 41.4 & 56.4 & 31.6 \\
\hline
\end{tabular*}
\caption{Ablation study on SciDisco training objectives. Performance is evaluated on DataSciBench. Success Rate reports pass rate, Completion Rate reports completion accuracy, and F1--F5 measure fine-grained performance across data-science sub-tasks. VLM was left empty because the backbone model lacks vision capability.}
\label{tab:ablation-results}
\end{table*}

\subsection{Models and Baselines}
\label{sec:models-baselines}

We organize baselines into three groups to separate general model capability from data-science-specific training. The proprietary group includes GPT-4o~\citep{openai2024gpt4o}, GPT-5 Mini~\citep{openai2025gpt5}, and Claude Sonnet 4~\citep{anthropic2025claude4}, representing strong closed-source agentic models. The open-source group includes DeepSeek-V4-Flash~\citep{deepseek2026v4flash}, Intern-S1-Pro~\citep{interns1pro2026}, and Qwen3-8B/14B/32B~\citep{yang2025qwen3}, allowing comparison against publicly available general models at different scales. The related-method group includes DeepAnalyze-8B~\citep{zhang2025deepanalyze}, AutoSDT-Coder-14B~\citep{li2025autosdt}, and DataMind-14B~\citep{qiao2025datamind}, which are specialized for data analysis or scientific discovery. SciDisco-14B is trained from Qwen3-14B using the SciDisco trajectory corpus and DiscoPO objective.

\subsection{Experimental Setup}
\label{sec:experimental-setup}

Starting from Qwen3-14B, we first apply cold-start SFT on SciDisco-produced verified trajectories and then perform DiscoPO agentic RL. Training uses the Slime post-training framework~\citep{slime2025}, with SGLang for rollout generation and inference~\citep{zheng2024sglang}. For the RL stage, we use a rollout batch size of 16 and a group size of 8. The resulting SciDisco-14B checkpoint is evaluated on the three benchmarks above. All training and inference are conducted on NVIDIA A100 GPUs; detailed hyperparameters, source licenses, and corpus distribution are reported in Appendix~\ref{app:env-details} and Appendix~\ref{app:training-details}.

\subsection{Main Results}
\label{sec:main-results}

Table~\ref{tab:main-results} compares proprietary, open-source, and specialised data-science agents across three scientific data-analysis benchmarks. SciDisco-14B obtains the strongest DiscoveryBench result among all listed models, reaching 35.2\% average HMS. This score exceeds the strongest proprietary baseline, GPT-5 Mini at 26.9\%, the strongest open-source baseline, Intern-S1-Pro at 34.7\%, and the strongest related-method baseline, DeepAnalyze-8B at 25.9\%.

SciDisco-14B also performs strongly on DataSciBench. It reaches 56.2\% Success Rate and 61.0\% Completion Rate, outperforming the related-method baselines on both metrics. The result remains below the strongest proprietary and open-source models on DataSciBench, but closes much of the gap while using a 14B open-weight backbone.

DABStep shows a different pattern. SciDisco-14B reaches 17.8\% average accuracy, below DeepAnalyze-8B at 38.9\%, Claude-Sonnet-4 at 29.7\%, and GPT-5 Mini at 23.8\%. The hard split remains challenging for all Qwen3-family models, with Qwen3-8B, Qwen3-14B, and Qwen3-32B all below 5\%. SciDisco improves over the Qwen3-14B backbone on DABStep, but does not match methods specialised for this benchmark regime.

\subsection{Ablation Studies}
\label{sec:ablation}

Table~\ref{tab:ablation-results} isolates the effect of the training objective on the 14B backbone using DataSciBench. SFT provides the main first-stage gain over Qwen3-14B, and adding GRPO gives a further but smaller improvement. The full SFT + DiscoPO setting obtains the strongest coarse-grained result, reaching 56.3\% Success Rate and 61.0\% Completion Rate.

The fine-grained metrics follow the same ordering. SFT + DiscoPO obtains the highest score on every reported sub-task, with the largest incremental gains over SFT + GRPO appearing in Data Modeling and Data Exploration. These two categories are the clearest ablation signal for the turn-level objective, while the detailed mechanism is analysed in Section~\ref{sec:analysis}.

\subsection{Analysis}
\label{sec:analysis}

The results support a specific reading of SciDisco's contribution. DiscoveryBench is closest to the paper's data-driven scientific discovery setting because it evaluates hypothesis construction over scientific datasets. SciDisco-14B's leading HMS score indicates that training on process-verifiable scientific environments improves the model's ability to form benchmark-matched scientific hypotheses from data.

DataSciBench provides a complementary executable analysis test. The gains over related methods and the ablation improvements in Table~\ref{tab:ablation-results} indicate that SciDisco also improves code-mediated data analysis under programmatic checks. The strongest ablation gains appear in Data Exploration and Data Modeling, the two categories most directly tied to intermediate analytical decisions.

DABStep exposes the boundary of the current training distribution. Its business data-analysis tasks often require completing operational analyses, reports, visualisations, or decision-support queries over enterprise-like tables. SciDisco instead trains on hypothesis-driven scientific environments where progress is defined by verifier-grounded evidence for statistical claims. The weaker DABStep result therefore suggests that the current environment distribution transfers better to scientific hypothesis analysis and executable scientific data analysis than to all forms of multi-step business analytics.

The ablation results locate where the training signal enters. SFT gives the largest first-stage gain by exposing the model to verifier-filtered executable trajectories. GRPO adds a smaller improvement over SFT, indicating that trajectory-level outcome optimisation supplies useful reinforcement. SFT + DiscoPO gives the strongest result because the reward is assigned to turns that complete hidden evidence-state transitions, rather than only to trajectories that end with a correct answer.

\section{Conclusion}
\label{sec:conclusion}
We presented SciDisco, a framework for training data-driven scientific discovery agents in process-verifiable environments. Hidden evidence DAGs and verifier checks make intermediate analytical progress trainable, while SciTh`eque compiles scientific datasets and hypothesis templates into sandboxed environments shared by DAG-grounded synthesis and DiscoPO. Evaluation results support SciDisco as a scalable pipeline for converting verified analytical evidence into process supervision.

\section*{Limitations}
SciDisco focuses on executable data-analysis workflows over text-readable scientific data, rather than wet-lab discovery, multimodal scientific artifacts, or open-ended hypothesis generation beyond the template space. The coverage of the environment distribution is also limited by the available source adapters, modality handlers, and hypothesis templates. Finally, environment-accepted progress is still a proxy for scientific quality: it verifies whether an analysis step was valid for the task contract, but it does not by itself establish that the agent made a novel scientific discovery.


\bibliography{custom}

@inproceedings{majumder2024discoverybench,
  title = {{DiscoveryBench}: Towards Data-Driven Discovery with Large Language Models},
  author = {Majumder, Bodhisattwa Prasad and Surana, Harshit and Agarwal, Dhruv and Dalvi Mishra, Bhavana and Meena, Abhijeetsingh and Prakhar, Aryan and Vora, Tirth and Khot, Tushar and Sabharwal, Ashish and Clark, Peter},
  booktitle = {The Thirteenth International Conference on Learning Representations},
  year = {2025},
  url = {https://openreview.net/forum?id=vyflgpwfJW}
}

@article{dabstep2025,
  title = {{DABstep}: Data Agent Benchmark for Multi-Step Reasoning},
  author = {Egg, Alex and Iglesias Goyanes, Martin and Kingma, Friso and Mora, Andreu and von Werra, Leandro and Wolf, Thomas},
  journal = {arXiv preprint arXiv:2506.23719},
  year = {2025},
  url = {https://arxiv.org/abs/2506.23719}
}

@article{zhang2025datascibench,
  title = {{DataSciBench}: An {LLM} Agent Benchmark for Data Science},
  author = {Zhang, Dan and Zhoubian, Sining and Cai, Min and Li, Fengzu and Yang, Lekang and Wang, Wei and Dong, Tianjiao and Hu, Ziniu and Tang, Jie and Yue, Yisong},
  journal = {arXiv preprint arXiv:2502.13897},
  year = {2025},
  url = {https://arxiv.org/abs/2502.13897}
}

@misc{openai2025gpt5,
  title = {Introducing {GPT-5} for Developers},
  author = {{OpenAI}},
  year = {2025},
  howpublished = {OpenAI Blog},
  url = {https://openai.com/index/introducing-gpt-5-for-developers/}
}

@misc{anthropic2025claude4,
  title = {Introducing {Claude 4}},
  author = {{Anthropic}},
  year = {2025},
  howpublished = {Anthropic News},
  url = {https://www.anthropic.com/news/claude-4}
}

@article{openai2024gpt4o,
  title = {{GPT-4o} System Card},
  author = {{OpenAI}},
  journal = {arXiv preprint arXiv:2410.21276},
  year = {2024},
  url = {https://arxiv.org/abs/2410.21276}
}

@misc{deepseek2026v4flash,
  title = {{DeepSeek-V4-Flash}},
  author = {{DeepSeek-AI}},
  year = {2026},
  howpublished = {Hugging Face Model Card},
  url = {https://huggingface.co/deepseek-ai/DeepSeek-V4-Flash}
}

@article{interns1pro2026,
  title = {{Intern-S1-Pro}: Scientific Multimodal Foundation Model at Trillion Scale},
  author = {{Intern-S1-Pro Team}},
  journal = {arXiv preprint arXiv:2603.25040},
  year = {2026},
  url = {https://arxiv.org/abs/2603.25040}
}

@article{yang2025qwen3,
  title = {{Qwen3} Technical Report},
  author = {Yang, An and Li, Anfeng and Yang, Baosong and others},
  journal = {arXiv preprint arXiv:2505.09388},
  year = {2025},
  url = {https://arxiv.org/abs/2505.09388}
}

@inproceedings{zheng2024sglang,
  title = {{SGLang}: Efficient Execution of Structured Language Model Programs},
  author = {Zheng, Lianmin and Yin, Liangsheng and Xie, Zhiqiang and Sun, Chuyue and Huang, Jeff and Yu, Cody Hao and Cao, Shiyi and Kozyrakis, Christos and Stoica, Ion and Gonzalez, Joseph E. and Barrett, Clark W. and Sheng, Ying},
  booktitle = {Advances in Neural Information Processing Systems},
  year = {2024},
  url = {https://arxiv.org/abs/2312.07104}
}

@misc{slime2025,
  title = {{slime}: An {SGLang}-Native Post-Training Framework for {RL} Scaling},
  author = {{THUDM}},
  year = {2025},
  howpublished = {Software repository},
  url = {https://github.com/THUDM/slime}
}

@article{zhang2025deepanalyze,
  title = {{DeepAnalyze}: Agentic Large Language Models for Autonomous Data Science},
  author = {Zhang, Shaolei and Fan, Ju and Fan, Meihao and Li, Guoliang and Du, Xiaoyong},
  journal = {arXiv preprint arXiv:2510.16872},
  year = {2025},
  url = {https://arxiv.org/abs/2510.16872}
}

@inproceedings{li2025autosdt,
  title = {{AutoSDT}: Scaling Data-Driven Discovery Tasks Toward Open Co-Scientists},
  author = {Li, Yifei and Moussa, Hanane Nour and Chen, Ziru and Chen, Shijie and Yu, Botao and Xue, Mingyi and Burns, Benjamin and Chiu, Tzu-Yao and Dey, Vishal and Lu, Zitong and Wei, Chen and Zhang, Qianheng and Zhang, Tianyu and Gao, Song and Huang, Xuhui and Ning, Xia and Ahmed, Nesreen K. and Payani, Ali and Sun, Huan},
  booktitle = {Proceedings of the 2025 Conference on Empirical Methods in Natural Language Processing},
  pages = {30396--30418},
  year = {2025},
  doi = {10.18653/v1/2025.emnlp-main.1546},
  url = {https://aclanthology.org/2025.emnlp-main.1546/}
}

@article{qiao2025datamind,
  title = {Scaling Generalist Data-Analytic Agents},
  author = {Qiao, Shuofei and Zhao, Yanqiu and Qiu, Zhisong and Wang, Xiaobin and Zhang, Jintian and Bin, Zhao and Zhang, Ningyu and Jiang, Yong and Xie, Pengjun and Huang, Fei and Chen, Huajun},
  journal = {arXiv preprint arXiv:2509.25084},
  year = {2025},
  url = {https://arxiv.org/abs/2509.25084}
}

@inproceedings{zhu2024statqa,
  title = {Are Large Language Models Good Statisticians?},
  author = {Zhu, Yizhang and Du, Shiyin and Li, Boyan and Luo, Yuyu and Tang, Nan},
  booktitle = {Advances in Neural Information Processing Systems},
  year = {2024},
  url = {https://arxiv.org/abs/2406.07815}
}

@inproceedings{liu2024qrdata,
  title = {Are {LLM}s Capable of Data-Based Statistical and Causal Reasoning? Benchmarking Advanced Quantitative Reasoning with Data},
  author = {Liu, Xiao and Wu, Zirui and Wu, Xueqing and Lu, Pan and Chang, Kai-Wei and Feng, Yansong},
  booktitle = {Findings of the Association for Computational Linguistics: ACL 2024},
  year = {2024},
  pages = {9215--9235},
  url = {https://aclanthology.org/2024.findings-acl.548/}
}

@inproceedings{gu2024blade,
  title = {{BLADE}: Benchmarking Language Model Agents for Data-Driven Science},
  author = {Gu, Ken and Shang, Ruoxi and Jiang, Ruien and Kuang, Keying and Lin, Richard-John and Lyu, Donghe and Mao, Yue and Pan, Youran and Wu, Teng and Yu, Jiaqian and Zhang, Yikun and Tianmai M. Zhang and Zhu, Lanyi and Merrill, Mike A. and Heer, Jeffrey and Althoff, Tim},
  booktitle = {Proceedings of the 2024 Conference on Empirical Methods in Natural Language Processing},
  year = {2024},
  url = {https://arxiv.org/abs/2408.09667}
}

@inproceedings{chen2025scienceagentbench,
  title = {{ScienceAgentBench}: Toward Rigorous Assessment of Language Agents for Data-Driven Scientific Discovery},
  author = {Chen, Ziru and Chen, Shijie and Ning, Yuting and Zhang, Qianheng and Wang, Boshi and Yu, Botao and Li, Yifei and Liao, Zeyi and Wei, Chen and Lu, Zitong and Dey, Vishal and Xue, Mingyi and Baker, Frazier N. and Burns, Benjamin and Adu-Ampratwum, Daniel and Huang, Xuhui and Ning, Xia and Gao, Song and Su, Yu and Sun, Huan},
  booktitle = {International Conference on Learning Representations},
  year = {2025},
  url = {https://openreview.net/forum?id=6z4YKr0GK6}
}

@article{abdulhai2023lmrlgym,
  title = {{LMRL} Gym: Benchmarks for Multi-Turn Reinforcement Learning with Language Models},
  author = {Abdulhai, Marwa and White, Isadora and Snell, Charlie and Sun, Charles and Hong, Joey and Zhai, Yuexiang and Xu, Kelvin and Levine, Sergey},
  journal = {arXiv preprint arXiv:2311.18232},
  year = {2023},
  url = {https://arxiv.org/abs/2311.18232}
}

@article{qian2025toolrl,
  title = {{ToolRL}: Reward is All Tool Learning Needs},
  author = {Qian, Cheng and Acikgoz, Emre Can and He, Qi and Wang, Hongru and Chen, Xiusi and Hakkani-Tur, Dilek and Tur, Gokhan and Ji, Heng},
  journal = {arXiv preprint arXiv:2504.13958},
  year = {2025},
  url = {https://arxiv.org/abs/2504.13958}
}

@inproceedings{liu2023agentbench,
  title = {{AgentBench}: Evaluating {LLM}s as Agents},
  author = {Liu, Xiao and Yu, Hao and Zhang, Hanchen and Xu, Yifan and Lei, Xuanyu and Lai, Hanyu and Gu, Yu and Ding, Hangliang and Men, Kaiwen and Yang, Kejuan and others},
  booktitle = {International Conference on Learning Representations},
  year = {2024},
  url = {https://arxiv.org/abs/2308.03688}
}

@article{chan2024mlebench,
  title = {{MLE}-bench: Evaluating Machine Learning Agents on Machine Learning Engineering},
  author = {Chan, Jun Shern and Chowdhury, Neil and Jaffe, Oliver and Aung, James and Sherburn, Dane and Mays, Evan and Starace, Giulio and Liu, Kevin and Maksin, Leon and Patwardhan, Tejal and Weng, Lilian and Madry, Aleksander},
  journal = {arXiv preprint arXiv:2410.07095},
  year = {2024},
  url = {https://arxiv.org/abs/2410.07095}
}

@article{wang2025ragen,
  title = {{RAGEN}: Understanding Self-Evolution in {LLM} Agents via Multi-Turn Reinforcement Learning},
  author = {Wang, Zihan and Wang, Kangrui and Wang, Qineng and Zhang, Pingyue and Li, Linjie and Yang, Zhengyuan and Jin, Xing and Yu, Kefan and Nguyen, Minh Nhat and Liu, Licheng and Gottlieb, Eli and Lu, Yiping and Cho, Kyunghyun and Wu, Jiajun and Fei-Fei, Li and Wang, Lijuan and Choi, Yejin and Li, Manling},
  journal = {arXiv preprint arXiv:2504.20073},
  year = {2025},
  url = {https://arxiv.org/abs/2504.20073}
}

@article{zeng2025turnlevelcredit,
  title = {Reinforcing Multi-Turn Reasoning in {LLM} Agents via Turn-Level Reward Design},
  author = {Wei, Quan and Zeng, Siliang and Li, Chenliang and Brown, William and Frunza, Oana and Deng, Wei and Schneider, Anderson and Nevmyvaka, Yuriy and Zhao, Yang Katie and Garcia, Alfredo and Hong, Mingyi},
  journal = {arXiv preprint arXiv:2505.11821},
  year = {2025},
  url = {https://arxiv.org/abs/2505.11821}
}

@article{li2026tlgrpo,
  title = {{TL-GRPO}: Turn-Level {RL} for Reasoning-Guided Iterative Optimization},
  author = {Li, Peiji and Li, Linyang and Sun, Handa and Mai, Wenjin and Chen, Yongkang and Li, Xiaozhe and Shen, Yue and Ma, Yichuan and Sun, Yiliu and Cao, Jiaxi and He, Zhishu and Wang, Bo and Zheng, Xiaoqing and Bi, Zhaori and Qiu, Xipeng and Guo, Qipeng and Chen, Kai and Lin, Dahua},
  journal = {arXiv preprint arXiv:2601.16480},
  year = {2026},
  url = {https://arxiv.org/abs/2601.16480}
}

@article{xie2026tips,
  title = {{TIPS}: Turn-Level Information-Potential Reward Shaping for Search-Augmented {LLM}s},
  author = {Xie, Yutao and Thomas, Nathaniel and Hansen, Nicklas and Fu, Yang and Li, Li Erran and Wang, Xiaolong},
  journal = {arXiv preprint arXiv:2603.22293},
  year = {2026},
  url = {https://arxiv.org/abs/2603.22293}
}

@inproceedings{guo2024dsagent,
  title = {{DS-Agent}: Automated Data Science by Empowering Large Language Models with Case-Based Reasoning},
  author = {Guo, Siyuan and Deng, Cheng and Wen, Ying and Chen, Hechang and Chang, Yi and Wang, Jun},
  booktitle = {International Conference on Machine Learning},
  year = {2024},
  url = {https://arxiv.org/abs/2402.17453}
}

@article{zhang2023datacopilot,
  title = {{Data-Copilot}: Bridging Billions of Data and Humans with Autonomous Workflow},
  author = {Zhang, Wenqi and Shen, Yongliang and Lu, Weiming and Zhuang, Yueting},
  journal = {arXiv preprint arXiv:2306.07209},
  year = {2023},
  url = {https://arxiv.org/abs/2306.07209}
}

@article{hong2024datainterpreter,
  title = {Data Interpreter: An {LLM} Agent for Data Science},
  author = {Hong, Sirui and Lin, Yizhang and Liu, Bang and Liu, Bangbang and Wu, Binhao and Zhang, Ceyao and Wei, Chenxing and Li, Danyang and Chen, Jiaqi and Zhang, Jiayi and Wang, Jinlin and Zhang, Li and Zhang, Lingyao and Yang, Min and Zhuge, Mingchen and Guo, Taicheng and Zhou, Tuo and Tao, Wei and Tang, Xiangru and Lu, Xiangtao and Zheng, Xiawu and Liang, Xinbing and Fei, Yaying and Cheng, Yuheng and Gou, Zhibin and Xu, Zongze and Wu, Chenglin},
  journal = {arXiv preprint arXiv:2402.18679},
  year = {2024},
  url = {https://arxiv.org/abs/2402.18679}
}

@article{li2024autokaggle,
  title = {{AutoKaggle}: A Multi-Agent Framework for Autonomous Data Science Competitions},
  author = {Li, Ziming and Zang, Qianbo and Ma, David and Guo, Jiawei and Zheng, Tuney and Liu, Minghao and Niu, Xinyao and Wang, Yue and Yang, Jian and Liu, Jiaheng and Zhong, Wanjun and Zhou, Wangchunshu and Huang, Wenhao and Zhang, Ge},
  journal = {arXiv preprint arXiv:2410.20424},
  year = {2024},
  url = {https://arxiv.org/abs/2410.20424}
}

@article{wang2022scienceworld,
  title = {{ScienceWorld}: Is Your Agent Smarter than a 5th Grader?},
  author = {Wang, Ruoyao and Jansen, Peter and Cote, Marc-Alexandre and Ammanabrolu, Prithviraj},
  journal = {arXiv preprint arXiv:2203.07540},
  year = {2022},
  url = {https://arxiv.org/abs/2203.07540}
}

@article{osullivan2024discoveryworld,
  title = {{DiscoveryWorld}: A Virtual Environment for Developing and Evaluating Automated Scientific Discovery Agents},
  author = {O'Sullivan, John and Lindsay, Alan and Magerko, Brian and Lieto, Antonio and Goel, Ashok K.},
  journal = {arXiv preprint arXiv:2406.06769},
  year = {2024},
  url = {https://arxiv.org/abs/2406.06769}
}

@inproceedings{zhou2023webarena,
  title = {{WebArena}: A Realistic Web Environment for Building Autonomous Agents},
  author = {Zhou, Shuyan and Xu, Frank F. and Zhu, Hao and Zhou, Xuhui and Lo, Robert and Sridhar, Abishek and Cheng, Xianyi and Bisk, Yonatan and Fried, Daniel and Alon, Uri and Neubig, Graham},
  booktitle = {International Conference on Learning Representations},
  year = {2024},
  url = {https://arxiv.org/abs/2307.13854}
}

@inproceedings{qiang2025mledojo,
  title = {{MLE-Dojo}: Interactive Environments for Empowering {LLM} Agents in Machine Learning Engineering},
  author = {Qiang, Rushi and Zhuang, Yuchen and Li, Yinghao and Dingu Sagar, V K and Zhang, Rongzhi and Li, Changhao and Wong, Ian Shu-Hei and Yang, Sherry and Liang, Percy and Zhang, Chao and Dai, Bo},
  booktitle = {Advances in Neural Information Processing Systems},
  year = {2025},
  url = {https://arxiv.org/abs/2505.07782}
}

@article{sandmle2026,
  title = {Synthetic Sandbox for Training Machine Learning Engineering Agents},
  author = {Zhou, Yuhang and Zhang, Lizhu and Wu, Yifan and Liu, Jiayi and Fan, Xiangjun and Zhao, Zhuokai and Yan, Hong},
  journal = {arXiv preprint arXiv:2604.04872},
  year = {2026},
  url = {https://arxiv.org/abs/2604.04872}
}

@article{schulman2017ppo,
  title = {Proximal Policy Optimization Algorithms},
  author = {Schulman, John and Wolski, Filip and Dhariwal, Prafulla and Radford, Alec and Klimov, Oleg},
  journal = {arXiv preprint arXiv:1707.06347},
  year = {2017},
  url = {https://arxiv.org/abs/1707.06347}
}

@article{shao2024deepseekmath,
  title = {{DeepSeekMath}: Pushing the Limits of Mathematical Reasoning in Open Language Models},
  author = {Shao, Zhihong and Wang, Peiyi and Zhu, Qihao and Xu, Runxin and Song, Junxiao and Bi, Xiao and Zhang, Haowei and Zhang, Mingchuan and Li, Y. K. and Wu, Y. and Guo, Daya},
  journal = {arXiv preprint arXiv:2402.03300},
  year = {2024},
  url = {https://arxiv.org/abs/2402.03300}
}

@article{guo2025deepseekr1,
  title = {{DeepSeek-R1}: Incentivizing Reasoning Capability in {LLM}s via Reinforcement Learning},
  author = {Guo, Daya and Yang, Dejian and Zhang, Haowei and Song, Junxiao and Zhang, Ruoyu and Xu, Runxin and Zhu, Qihao and Ma, Shirong and Wang, Peiyi and Bi, Xiao and others},
  journal = {arXiv preprint arXiv:2501.12948},
  year = {2025},
  url = {https://arxiv.org/abs/2501.12948}
}

@article{yu2025dapo,
  title = {{DAPO}: An Open-Source {LLM} Reinforcement Learning System at Scale},
  author = {Yu, Qiying and Zhang, Zheng and Zhu, Ruofei and Yuan, Yufeng and Zuo, Xiaochen and Yue, Yu and Fan, Tiantian and Liu, Gaohong and Liu, Lingjun and Liu, Xin and others},
  journal = {arXiv preprint arXiv:2503.14476},
  year = {2025},
  url = {https://arxiv.org/abs/2503.14476}
}

@article{liu2025drgrpo,
  title = {Understanding {R1-Zero}-Like Training: A Critical Perspective},
  author = {Liu, Zichen and Liu, Chang and Liu, Xueqing and Guo, Daya and Chen, Jun and Zhou, Chunting},
  journal = {arXiv preprint arXiv:2503.20783},
  year = {2025},
  url = {https://arxiv.org/abs/2503.20783}
}

@article{srpo2026,
  title = {Unifying Group-Relative and Self-Distillation Policy Optimization via Sample Routing},
  author = {Li, Gengsheng and Yang, Tianyu and Fang, Junfeng and Song, Mingyang and Zheng, Mao and Guo, Haiyun and Zhang, Dan and Wang, Jinqiao and Chua, Tat-Seng},
  journal = {arXiv preprint arXiv:2604.02288},
  year = {2026},
  url = {https://arxiv.org/abs/2604.02288}
}

@article{feng2025gigpo,
  title = {Group-in-Group Policy Optimization for {LLM} Agent Training},
  author = {Feng, Lang and Xue, Zhenghai and Liu, Tingcong and An, Bo},
  journal = {arXiv preprint arXiv:2505.10978},
  year = {2025},
  url = {https://arxiv.org/abs/2505.10978}
}

@article{guo2025spo,
  title = {Segment Policy Optimization: Effective Segment-Level Credit Assignment in {RL} for Large Language Models},
  author = {Guo, Yiran and Xu, Lijie and Liu, Jie and Ye, Dan and Qiu, Shuang},
  journal = {arXiv preprint arXiv:2505.23564},
  year = {2025},
  url = {https://arxiv.org/abs/2505.23564}
}

@article{lu2026hisr,
  title = {{HISR}: Hindsight Information Modulated Segmental Process Rewards for Multi-Turn Agentic Reinforcement Learning},
  author = {Lu, Zhicong and Lin, Zichuan and Jia, Wei and Tian, Changyuan and Ye, Deheng and Li, Peiguang and Jin, Li and Liu, Nayu and Xu, Guangluan and Feng, Wei},
  journal = {arXiv preprint arXiv:2603.18683},
  year = {2026},
  url = {https://arxiv.org/abs/2603.18683}
}

@inproceedings{wang2025steca,
  title = {{STeCa}: Step-Level Trajectory Calibration for {LLM} Agent Learning},
  author = {Wang, Hanlin and Wang, Jian and Leong, Chak Tou and Li, Wenjie},
  booktitle = {Findings of the Association for Computational Linguistics: ACL 2025},
  year = {2025},
  url = {https://arxiv.org/abs/2502.14276}
}

@article{wang2025stepsearch,
  title = {{StepSearch}: Igniting {LLM}s Search Ability via Step-Wise Proximal Policy Optimization},
  author = {Wang, Ziliang and Zheng, Xuhui and An, Kang and Ouyang, Cijun and Cai, Jialu and Wang, Yuhang and Wu, Yichao},
  journal = {arXiv preprint arXiv:2505.15107},
  year = {2025},
  url = {https://arxiv.org/abs/2505.15107}
}

@article{zong2026at2po,
  title = {{AT$^2$PO}: Agentic Turn-Based Policy Optimization via Tree Search},
  author = {Zong, Zefang and Chen, Dingwei and Li, Yang and Yi, Qi and Zhou, Bo and Li, Chengming and Qian, Bo and Chen, Peng and Jiang, Jie},
  journal = {arXiv preprint arXiv:2601.04767},
  year = {2026},
  url = {https://arxiv.org/abs/2601.04767}
}

@misc{uci2023repository,
  title = {The {UCI} Machine Learning Repository},
  author = {Kelly, Markelle and Longjohn, Rachel and Nottingham, Kolby},
  year = {2023},
  howpublished = {Data repository},
  url = {https://archive.ics.uci.edu}
}

@article{vanschoren2013openml,
  title = {{OpenML}: Networked Science in Machine Learning},
  author = {Vanschoren, Joaquin and van Rijn, Jan N. and Bischl, Bernd and Torgo, Luis},
  journal = {SIGKDD Explorations},
  volume = {15},
  number = {2},
  pages = {49--60},
  year = {2013},
  doi = {10.1145/2641190.2641198},
  url = {https://doi.org/10.1145/2641190.2641198}
}

@misc{fred2026,
  title = {{FRED}, Federal Reserve Economic Data},
  author = {{Federal Reserve Bank of St. Louis}},
  year = {2026},
  howpublished = {Data service},
  note = {Accessed 25 May 2026},
  url = {https://fred.stlouisfed.org/}
}

@article{uniprot2025,
  title = {{UniProt}: The Universal Protein Knowledgebase in 2025},
  author = {{The UniProt Consortium}},
  journal = {Nucleic Acids Research},
  volume = {53},
  number = {D1},
  pages = {D609--D617},
  year = {2025},
  doi = {10.1093/nar/gkae1010},
  url = {https://academic.oup.com/nar/article/53/D1/D609/7902999}
}

@misc{cdcfluview2026,
  title = {{FluView} Interactive},
  author = {{Centers for Disease Control and Prevention}},
  year = {2026},
  howpublished = {Public health data service},
  note = {Accessed 25 May 2026},
  url = {https://www.cdc.gov/fluview/overview/fluview-interactive.html}
}

@misc{cdcplaces2026,
  title = {{PLACES}: Local Data for Better Health},
  author = {{Centers for Disease Control and Prevention}},
  year = {2026},
  howpublished = {Public health data service},
  note = {Accessed 25 May 2026},
  url = {https://www.cdc.gov/places/}
}

@misc{usgs2026,
  title = {{USGS} Water Data for the Nation},
  author = {{U.S. Geological Survey}},
  year = {2026},
  howpublished = {Earth-science data service},
  note = {Accessed 25 May 2026},
  url = {https://waterdata.usgs.gov/nwis}
}

@inproceedings{godahewa2021monash,
  title = {Monash Time Series Forecasting Archive},
  author = {Godahewa, Rakshitha and Bergmeir, Christoph and Webb, Geoffrey I. and Hyndman, Rob J. and Montero-Manso, Pablo},
  booktitle = {Advances in Neural Information Processing Systems Datasets and Benchmarks Track},
  year = {2021},
  url = {https://arxiv.org/abs/2105.06643}
}

@article{szklarczyk2023string,
  title = {The {STRING} Database in 2023: Protein-Protein Association Networks and Functional Enrichment Analyses for Any Sequenced Genome of Interest},
  author = {Szklarczyk, Damian and Kirsch, Rebecca and Koutrouli, Mikaela and Nastou, Katerina and Mehryary, Farrokh and Hachilif, Rayan and Gable, Annika L. and Fang, Tao and Doncheva, Nadezhda T. and Pyysalo, Sampo and Bork, Peer and Jensen, Lars J. and von Mering, Christian},
  journal = {Nucleic Acids Research},
  volume = {51},
  number = {D1},
  pages = {D638--D646},
  year = {2023},
  doi = {10.1093/nar/gkac1000},
  url = {https://academic.oup.com/nar/article/51/D1/D638/6825341}
}

@inproceedings{hu2020ogb,
  title = {Open Graph Benchmark: Datasets for Machine Learning on Graphs},
  author = {Hu, Weihua and Fey, Matthias and Zitnik, Marinka and Dong, Yuxiao and Ren, Hongyu and Liu, Bowen and Catasta, Michele and Leskovec, Jure},
  booktitle = {Advances in Neural Information Processing Systems},
  year = {2020},
  url = {https://arxiv.org/abs/2005.00687}
}

@misc{yelp2026dataset,
  title = {Yelp Open Dataset},
  author = {{Yelp Inc.}},
  year = {2026},
  howpublished = {Open dataset},
  note = {Accessed 25 May 2026},
  url = {https://business.yelp.com/data/resources/open-dataset/}
}

\clearpage
\appendix

\section{Task Environment and Data Coverage}
\label{app:env-details}

SciTh\`eque is a collection of self-contained scientific discovery environments. Each environment gives the agent a natural-language hypothesis, task data, and an interactive analysis session. The context shown to the agent contains only the task description and data. The hidden materials used for verification are kept outside the policy context and are used only by the environment to judge progress and final correctness.

\noindent\textbf{Environment snapshot.} This environment snapshot contains 1,686 materialised environments across five modalities. Table~\ref{tab:app-modality-counts} reports the modality distribution. The corpus is not limited to standard tabular prediction: it also includes time-series, spatial, sequence, and graph settings so that the trained agent must reuse the same scientific-interaction protocol across heterogeneous data forms.

\begin{table}[!htbp]
\centering
\small
\setlength{\tabcolsep}{5pt}
\begin{tabular}{lrr}
\hline
\textbf{Modality} & \textbf{Environments} & \textbf{Share} \\
\hline
Tabular & 1,024 & 60.7\% \\
Time series & 287 & 17.0\% \\
Spatial & 206 & 12.2\% \\
Sequence & 114 & 6.8\% \\
Graph & 55 & 3.3\% \\
\hline
Total & 1,686 & 100.0\% \\
\hline
\end{tabular}
\caption{Materialized Environment Distribution by Modality}
\label{tab:app-modality-counts}
\end{table}

\noindent\textbf{Outcome and domain distribution.} The corpus spans continuous, binary, count, multiclass, and other structured outcome types. Table~\ref{tab:app-outcome-counts} reports the outcome-kind distribution, with ordinal, modality-specific, and survival outcomes grouped as other structured outcomes. Table~\ref{tab:app-domain-counts} reports the domain distribution, led by economics, medicine, biology, and earth science. The aggregate other-domain category contains additional scientific areas with distinct analysis conventions.

\begin{table}[!htbp]
\centering
\small
\setlength{\tabcolsep}{5pt}
\begin{tabular}{lrr}
\hline
\textbf{Outcome kind} & \textbf{Environments} & \textbf{Share} \\
\hline
Continuous & 716 & 42.5\% \\
Binary & 511 & 30.3\% \\
Count & 213 & 12.6\% \\
Multiclass & 209 & 12.4\% \\
Other structured & 37 & 2.2\% \\
\hline
Total & 1,686 & 100.0\% \\
\hline
\end{tabular}
\caption{Environment Distribution by Outcome Kind}
\label{tab:app-outcome-counts}
\end{table}

\begin{table}[!htbp]
\centering
\scriptsize
\setlength{\tabcolsep}{2pt}
\begin{tabular}{@{}p{0.31\columnwidth}@{\hspace{0.35em}}r@{\hspace{0.9em}}p{0.31\columnwidth}@{\hspace{0.35em}}r@{}}
\hline
\textbf{Domain} & \textbf{N} & \textbf{Domain} & \textbf{N} \\
\hline
Economics & 452 & Medicine & 334 \\
Biology & 325 & Earth science & 219 \\
Engineering & 127 & Epidemiology & 119 \\
Other scientific domains & 110 & & \\
\hline
\end{tabular}
\caption{Environment Distribution by Scientific Domain}
\label{tab:app-domain-counts}
\end{table}

\noindent\textbf{Stress variants.} The environment snapshot includes 504 stress variants derived from parent hypotheses. They perturb task settings while preserving the same interaction protocol and are counted separately from the aggregate corpus totals.

\noindent\textbf{Data sources and licences.} The environments are drawn from a mixture of general machine-learning repositories, open scientific time-series sources, biological databases, spatial and epidemiological data sources, economic indicators, and graph benchmarks~\citep{uci2023repository,vanschoren2013openml,fred2026,uniprot2025,cdcfluview2026,cdcplaces2026,usgs2026,yelp2026dataset,godahewa2021monash,szklarczyk2023string,hu2020ogb}. Table~\ref{tab:app-source-counts} reports the source distribution and the licence or access metadata recorded in the materialised catalogue. OpenML is listed with multiple licences because the source records licence metadata per dataset.

\begin{table*}[!t]
\centering
\scriptsize
\setlength{\tabcolsep}{4pt}
\begin{tabular}{p{0.18\linewidth}r p{0.34\linewidth}p{0.30\linewidth}}
\hline
\textbf{Source} & \textbf{N} & \textbf{Recorded licence/access basis} & \textbf{Metadata scope} \\
\hline
UCI & 678 & \texttt{uci-redistributable} & Source-level catalogue label \\
OpenML & 475 & \texttt{public-domain}, \texttt{cc0-1.0}, \texttt{cc-by-4.0}, \texttt{cc-by-sa-4.0} & Dataset-level OpenML metadata \\
FRED & 151 & \texttt{public-domain} & Source-level catalogue label \\
UniProt & 114 & \texttt{cc-by-4.0} & Source-level catalogue label \\
CDC FluView & 65 & \texttt{public-domain} & Source-level catalogue label \\
USGS & 41 & \texttt{public-domain} & Source-level catalogue label \\
CDC PLACES & 36 & \texttt{public-domain} & Source-level catalogue label \\
Yelp & 40 & \texttt{cc-by-nc-nd-4.0} & Source-level catalogue label \\
Monash & 31 & \texttt{cc-by-4.0} & Source-level catalogue label \\
STRING yeast & 34 & \texttt{cc-by-4.0} & Source-level catalogue label \\
OGB & 21 & \texttt{mit} & Source-level catalogue label \\
\hline
\end{tabular}
\caption{Environment Counts and Metadata Scope by Data Source}
\label{tab:app-source-counts}
\end{table*}

\noindent\textbf{Intended use.} The SciTh\`eque environments, derived trajectories, and trained checkpoints are intended for research on training and evaluating scientific-discovery agents. They are not validated scientific findings, medical or public-health advice, economic recommendations, or operational decision-support systems. Derived artifacts retain the access conditions of their upstream sources: public-domain and permissively licensed sources support redistribution only within their stated terms, while sources with non-commercial or no-derivatives conditions are used as research inputs and are not relicensed for broader downstream deployment.

\noindent\textbf{Privacy and content screening.} The corpus is built from open scientific, public-service, and benchmark-style data sources rather than newly collected private communications. Source adapters materialise structured analysis variables and exclude direct identity fields when those fields are not needed for the scientific task. For example, the Yelp adapter keeps geographic coordinates, open-status, rating, review-count, and structured business attributes, but does not materialise business names, user identifiers, or review text. This screening is schema- and adapter-based, so it should not be read as a guarantee that every upstream source is free of sensitive or offensive content. The released task prompts and training traces expose only task data and runtime observations; hidden verifier files, reference answers, and tolerance policies are kept outside the policy context and are separately checked by the leakage audit in Appendix~\ref{app:dag-acceptor-details}.

\noindent\textbf{Data provenance and reproducibility.} These counts are snapshot metadata for the environment corpus. Each task stores its source adapter, source dataset identifier, upstream provenance and licence metadata, modality, analysis template, prompt, hidden evidence graph, final checker, and data checksums.

\noindent\textbf{Benchmark-overlap audit.} The training corpus is audited against the evaluation snapshots used in Section~\ref{sec:benchmarks}. The audit compares SHA-256 hashes of materialised training data files with benchmark data files, then separately compares normalised source identifiers, task identifiers, benchmark directory names, and file stems. The audit covers 2,461 training data files with 991 unique file hashes and finds no direct overlap with the available benchmark snapshots: 0/165 DiscoveryBench files, 0/7 DABStep files, and 0/548 DataSciBench files. It also finds zero normalised source-identifier matches for all three benchmarks. The audit is scoped to exact file reuse and source-ID reuse, while broad topical similarity is expected across independently sourced scientific datasets.

\noindent\textbf{Environment contract.} Each environment separates three roles. The prompt specifies the scientific question and exposes the data. The hidden progress checker defines the ordered evidence needed before a final answer is admissible. The final-answer checker evaluates the submitted scientific claims against task-local reference values and tolerances. The checker is compiled from the same accepted reference analysis that materialises the task, while the tolerances and reference values remain outside the prompt and training transcript. This separation makes intermediate progress inspectable while preserving task-local final scoring.

\noindent\textbf{Example task.} A time-series environment can be written as \(u_j=(h_j,d_j,v_j)\). Here \(h_j\) asks whether an age-specific influenza-like-illness count helps predict total influenza-like-illness counts at a fixed weekly lag. The data component \(d_j\) is a weekly regional public-health series with a time index, a target count, age-specific count variables, patient totals, provider totals, and percentage-based illness measures. The verifier \(v_j\) checks that the trajectory inspects the series, assesses stationarity and sample size, fits an appropriate lagged time-series analysis, runs residual and stability diagnostics, performs a sensitivity check, and submits a conclusion whose significance bucket, effect-strength category, uncertainty verdict, and diagnostic outcomes match the task-local reference within the declared tolerances.

\section{Verifiability and Leakage Controls}
\label{app:dag-acceptor-details}

The environment is designed so that intermediate reward comes from scientific state transitions rather than from matching hidden answer strings. A hidden evidence graph specifies which analysis steps are currently eligible, what observable evidence they must produce, and which downstream claims they support. The policy never sees this graph directly. Table~\ref{tab:app-visibility-split} separates what is visible to the agent from what is used only by the verifier.

\begin{table}[!htbp]
\centering
\small
\setlength{\tabcolsep}{3pt}
\begin{tabular}{p{0.24\linewidth}p{0.32\linewidth}p{0.34\linewidth}}
\hline
\textbf{Surface} & \textbf{Visible to the agent} & \textbf{Used by the verifier} \\
\hline
Task prompt & Hypothesis, available data, requested answer format, and analysis constraints. & Prompt consistency checks and final-answer field expectations. \\
Data files & Input data available to the agent for analysis. & Runtime evidence produced by the agent from those data. \\
Intermediate observations & Execution output, errors, summaries, and feedback after each action. & Whether the action produced new evidence for exactly one currently eligible step. \\
Hidden evidence graph & Not visible. & Prerequisites, required evidence, supported claims, and progress state. \\
Reference answers & Not visible. & Final-answer scoring and task-local tolerance checks. \\
\hline
\end{tabular}
\caption{Visibility split between agent context and verifier state}
\label{tab:app-visibility-split}
\end{table}

\noindent\textbf{Evidence graph record.} Each hidden evidence graph is stored as a task-local record with fields for \texttt{task\_id}, \texttt{node\_id}, process primitive, prerequisite node set, admissible operations, required evidence fields, supported derived claims, tolerance policy, hidden reference values, and visibility flag. Table~\ref{tab:app-rollout-dag} shows the hidden DAG for the same CDC FluView task whose prompt and accepted rollout appear in Tables~\ref{tab:app-user-prompt} and~\ref{tab:app-real-rollout-example}. Node identifiers are shown here for auditability only; they are not included in the prompt or transcript.

\begin{table*}[!t]
\centering
\scriptsize
\setlength{\tabcolsep}{4pt}
\begin{tabular}{p{0.24\linewidth}p{0.13\linewidth}p{0.12\linewidth}p{0.32\linewidth}c}
\hline
\textbf{Hidden node} & \textbf{Primitive} & \textbf{Prerequisite} & \textbf{Verifier evidence} & \textbf{Rollout turn} \\
\hline
\texttt{setup\_time\_series\_data} & Data setup & None & Loaded frame, role map, target, candidate variable, time index, and frequency. & 1 \\
\texttt{prepare\_series\_frame} & EDA summary & Setup & Sorted series frame with sample-size, shape, missingness, stationarity, and serial-dependence metrics. & 2 \\
\texttt{build\_lagged\_design} & Feature extraction & Series frame & Lag-1 design matrix with target, candidate, control, and lagged columns. & 3 \\
\texttt{fit\_temporal\_model} & Model fit & Lagged design & P-value and effect-size evidence for the selected temporal model. & 4 \\
\texttt{check\_stationarity} & Diagnostic & Model fit & ADF stationarity evidence for the target series. & 5 \\
\texttt{check\_residual\_whiteness} & Diagnostic & Model fit & Ljung-Box and lag-1 autocorrelation evidence for residual dependence. & 6 \\
\texttt{check\_arch\_effects} & Diagnostic & Model fit & ARCH-effect diagnostic over fitted residuals. & 7 \\
\texttt{estimate\_temporal\_ci} & Uncertainty & Model fit & Confidence interval evidence for whether the effect overlaps the null. & 8 \\
\texttt{check\_convergence} & Diagnostic & Model fit & Warning-aware convergence status for a fitted or alternative temporal model. & 9 \\
\texttt{check\_sensitivity} & Robustness & Model fit & Main and alternative verdict comparison for method sensitivity. & 10 \\
\texttt{submit} & Terminal & Nodes 2--10 & Submitted claims pass the final checker after required evidence is complete. & 11 \\
\hline
\end{tabular}
\caption{Hidden evidence DAG for a CDC FluView time-series task}
\label{tab:app-rollout-dag}
\end{table*}

\noindent\textbf{Progress rule.} A turn receives progress only when it produces new, inspectable evidence for exactly one eligible scientific step. Eligibility is defined by the current frontier of the hidden graph: a node is available only after all of its prerequisites have been accepted. Repeating an already completed analysis, jumping to a future step before its prerequisites, bundling several steps into one action, or submitting the final answer prematurely leaves the hidden progress state unchanged. The terminal submission is credited only after the required intermediate evidence has been completed and the submitted claims pass the final checker.

\noindent\textbf{Claim derivation.} For claim-bearing steps, the verifier derives the relevant quantities from runtime evidence rather than trusting printed final literals. This prevents a trajectory from receiving progress simply by echoing a target value or by writing an answer-shaped string. A step must therefore both execute the appropriate analysis and expose enough structured evidence for the environment to validate the scientific quantity being claimed. Final checkers compare submitted answer fields against task-local references using hidden tolerance policies; the tolerance values are never serialized into prompts or exported SFT transcripts.

\noindent\textbf{Expert review.} The verifier mechanism was manually reviewed by 10 domain experts whose expertise covered the scientific domains represented in the environment corpus. The review focused on whether the evidence primitives, prerequisite ordering, diagnostic requirements, and acceptance criteria were appropriate for the corresponding domain and analysis template. Reviewers inspected the mechanism-level verifier specifications rather than model outputs, so this audit checks the scientific plausibility of the progress rules rather than measuring rollout-level verifier precision.

\noindent\textbf{Leakage controls.} Prompts and training traces are screened for direct exposure of reference answers, hidden step identifiers, hidden checklist language, task-local tolerance values, or checker code. The final-answer checker and hidden progress specification are not copied into the training transcript. The materialisation audit scans prompts before manifest emission, and the transcript-conversion audit scans retained SFT traces before training-data export. A failed audit excludes the task or trace from the emitted corpus. These checks are especially important for supervised trajectories: a retained trace must demonstrate the analysis path, not merely reproduce values that were available to the trajectory generator.

\noindent\textbf{Example verifier trace.} Table~\ref{tab:app-verifier-trace} shows the verifier information associated with the accepted 10-step CDC FluView rollout in Table~\ref{tab:app-real-rollout-example}. The table reports the hidden verifier match and the visible evidence exposed to the policy after each action. During training, the policy observes the accepted flag and the observation text; the node labels are used only by the environment and are shown here for auditability.

\begin{table*}[!htbp]
\centering
\scriptsize
\setlength{\tabcolsep}{3pt}
\begin{tabular}{r p{0.23\linewidth} p{0.24\linewidth} p{0.36\linewidth}}
\hline
\textbf{Turn} & \textbf{Verifier match} & \textbf{Decision} & \textbf{Visible evidence} \\
\hline
1 & \texttt{setup\_time\_series\_data} & Accepted setup & Loaded 1,379 rows and 12 columns; printed the first rows. \\
2 & \texttt{prepare\_series\_frame} & Accepted EDA summary & Reported rows, effective sample size, column count, and descriptive statistics. \\
3 & \texttt{build\_lagged\_design} & Accepted feature extraction & Built a lag-1 design matrix with 1,378 rows and seven columns. \\
4 & \texttt{fit\_temporal\_model} & Accepted model fit & Reported temporal model evidence with \(p=9.38\times10^{-26}\). \\
5 & \texttt{check\_stationarity} & Accepted diagnostic & Reported ADF stationarity evidence with \(p=1.41\times10^{-4}\). \\
6 & \texttt{check\_residual\_whiteness} & Accepted diagnostic & Reported Ljung--Box residual evidence and lag-1 autocorrelation. \\
7 & \texttt{check\_arch\_effects} & Accepted diagnostic & Reported ARCH-effect evidence with \(p=1.24\times10^{-266}\). \\
8 & \texttt{estimate\_temporal\_ci} & Accepted uncertainty check & Reported the effect confidence interval, 5.25--5.41. \\
9 & \texttt{check\_convergence} & Accepted diagnostic & Reported warning-aware convergence status as yes. \\
10 & \texttt{check\_sensitivity} & Accepted robustness check & Reported that the main verdict was not method-sensitive across three alternatives. \\
11 & \texttt{submit} & Accepted final submission & Final submission accepted after the required verifier state was complete. \\
\hline
\end{tabular}
\caption{Verifier trace for the CDC FluView rollout}
\label{tab:app-verifier-trace}
\end{table*}

\noindent\textbf{Leakage exposure scan.} The leakage scan checks the final RL prompt manifest and retained SFT traces for direct exposure of hidden verifier state outside the runtime evidence produced by executed analysis. The scan in Table~\ref{tab:app-leakage-audit} covers reference answer values in prompts or answer-format hints, task-local tolerance strings, hidden node identifiers, checker or gold-artifact markers, and generator-only markers. 

\begin{table*}[!htbp]
\centering
\small
\setlength{\tabcolsep}{5pt}
\begin{tabular}{llrr}
\hline
\textbf{Surface} & \textbf{Audit target} & \textbf{Checked} & \textbf{Flagged} \\
\hline
Prompt & Reference answer values & 1,686 & 0 \\
Prompt & Tolerance strings & 1,686 & 0 \\
Prompt & Hidden node identifiers & 1,686 & 0 \\
Prompt & Checker or gold markers & 1,686 & 0 \\
Prompt & Generator-only markers & 1,686 & 0 \\
SFT trace & Hidden node identifiers & 5,620 & 0 \\
SFT trace & Tolerance strings & 5,620 & 0 \\
SFT trace & Checker or gold markers & 5,620 & 0 \\
SFT trace & Generator-only markers & 5,620 & 0 \\
\hline
\end{tabular}
\caption{Leakage exposure scan for prompts and retained SFT traces}
\label{tab:app-leakage-audit}
\end{table*}

\section{Trajectory Synthesis and Filtering}
\label{app:sft-details}

The cold-start corpus is built from successful environment interactions rather than from free-form teacher answers. Each retained trajectory must complete the hidden evidence graph, receive accepted progress at every required analysis step, and pass the final-answer checker. Table~\ref{tab:app-sft-stages} summarizes the synthesis and filtering stages.

\begin{table}[!htbp]
\centering
\small
\setlength{\tabcolsep}{3pt}
\begin{tabular}{p{0.20\linewidth}p{0.34\linewidth}p{0.34\linewidth}}
\hline
\textbf{Stage} & \textbf{Purpose} & \textbf{Admission criterion} \\
\hline
Evidence planning & Decompose the task into a prerequisite-ordered sequence of scientific analysis steps. & The hidden graph is well formed and all required claims are represented. \\
Trajectory scheduling & Create valid analysis orders or presentation variants for each environment. & The schedule respects prerequisites and ends with a final submission. \\
Execution & Run each scheduled analysis step in an isolated environment and collect observations. & The action executes within the turn budget and produces observable evidence. \\
Verification & Check each turn against the hidden progress rule and score the final answer. & Every required analysis step is accepted and the final answer passes. \\
Conversion & Keep only messages suitable for model training. & The transcript fits the token budget and passes leakage checks for hidden labels, hidden tolerances, reference values, and answer strings. \\
\hline
\end{tabular}
\caption{Trajectory Synthesis and Filtering Pipeline for the Cold-Start Corpus}
\label{tab:app-sft-stages}
\end{table}

\noindent\textbf{Retained corpus.} The cold-start corpus contains 5,620 verified trajectories produced from accepted environment interactions.

\noindent\textbf{Filtering boundary.} Filtered traces are kept for diagnostics but excluded from training. The filter enforces whether a trajectory demonstrates the required scientific process rather than only whether it reaches the final answer.

\section{Training Details}
\label{app:training-details}

\begin{table}[!htbp]
\centering
\small
\setlength{\tabcolsep}{4pt}
\begin{tabular}{p{0.43\columnwidth}p{0.48\columnwidth}}
\hline
\textbf{Hyperparameter} & \textbf{Value} \\
\hline
\multicolumn{2}{l}{\textit{Cold-start supervised tuning}} \\
Backbone & Qwen3-14B \\
Corpus & SciDisco-produced \\
Epochs & 1 \\
Effective batch size & 128 \\
Learning rate & $1\times10^{-5}$ \\
Min learning rate & $1\times10^{-6}$ \\
Schedule & Cosine decay \\
Warm-up fraction & 0.1 \\
Weight decay & 0.1 \\
Context length & 24,576 tokens \\
\hline
\multicolumn{2}{l}{\textit{Agentic RL}} \\
Initial checkpoint & SciDisco-14B-SFT \\
Learning rate & \(1\times10^{-6}\) \\
Schedule & Constant \\
Rollout batch size & 16 \\
Global batch size & 128 \\
Rollout group size & 8 \\
Max turns & 30 \\
Per-turn generation cap & 4,096 tokens \\
Prompt cap & 8,192 tokens \\
Trajectory context cap & 24,576 tokens \\
Sampling temperature & 1.0 \\
Top-p & 0.95 \\
Clip high & 0.28 \\
Clip low & 0.2 \\
Weight decay & 0.1 \\
Adam beta 1 & 0.9 \\
Adam beta 2 & 0.98 \\
Action timeout & 60 s \\
\hline
\end{tabular}
\caption{Training and Rollout Hyperparameters}
\label{tab:app-training-settings}
\end{table}

Training has two stages. Cold-start supervised tuning teaches the solver the interaction format and rhythm of evidence-backed analysis. Agentic RL then optimizes the same protocol using environment turn-level progress signals. Table~\ref{tab:app-training-settings} reports the training and rollout hyperparameters.

\noindent\textbf{Turn-level signal.} For each prompt group, the environment records turns that complete new scientific progress. A rollout receives positive signal on turns that advance the hidden evidence state; invalid, repeated, premature, or bundled turns receive no progress. The RL objective compares sibling rollouts at the same interaction position, emphasizing which action was better under the same prompt and comparable history length.

\noindent\textbf{Filtering of uninformative groups.} Prompt groups with no turn-level contrast are removed before optimization. This avoids updates where all siblings make identical progress, all fail, or all terminate without a distinguishable learning signal. The filter is therefore not a data-quality heuristic; it removes batches whose relative advantage is mathematically uninformative.

\noindent\textbf{Execution budget.} Each scientific analysis action runs under a fixed wall-clock timeout. The timeout bounds inefficient code and accidental infinite computations while allowing common statistical, graph, time-series, and spatial analyses to complete.

\section{DiscoPO Objective Details}
\label{app:discopo-objective-details}

The policy update assigns $A_{j,k,t}$ to the policy-generated tokens in turn $t$ and excludes observation tokens from the loss. Let $\mathcal{I}_{j,k,t}$ denote the generated-token positions in that turn. For $i\in\mathcal{I}_{j,k,t}$, let $y_i$ be the generated token and $p_i$ its prefix. The likelihood ratio is:
\begin{equation}
\rho_i(\theta)
=
\frac{\pi_{\theta}(y_i\mid p_i)}
{\pi_{\mathrm{old}}(y_i\mid p_i)}.
\label{eq:discopo-ratio}
\end{equation}
The clipped ratio is:
\begin{equation}
\bar{\rho}_i(\theta)
=
\mathrm{clip}
\left(
\rho_i(\theta),
1-\epsilon_{\mathrm{low}},
1+\epsilon_{\mathrm{high}}
\right).
\label{eq:discopo-clip}
\end{equation}
For each generated token, the clipped surrogate is:
\begin{equation}
\ell_i(\theta)
=
\min\left(
\rho_i(\theta)A_{j,k,t},
\bar{\rho}_i(\theta)A_{j,k,t}
\right).
\label{eq:discopo-token-loss}
\end{equation}
DiscoPO optimizes the turn-averaged objective:
\begin{equation}
\begin{array}{c}
\bar{\ell}_{j,k,t}(\theta)
=
\frac{1}{|\mathcal{I}_{j,k,t}|}
\sum\limits_{i\in\mathcal{I}_{j,k,t}}
\ell_i(\theta),\\[0.5em]
\mathcal{L}_{\mathrm{DiscoPO}}(\theta)
=
-\mathrm{E}_{j,k,t}\left[
\bar{\ell}_{j,k,t}(\theta)
\right].
\end{array}
\label{eq:discopo-objective-expanded}
\end{equation}

\section{Prompt and Rollout Trace}
\label{app:prompt-rollout-example}

The interaction trace consists of a system prompt, a task prompt, alternating assistant actions and environment observations, and a terminal submission. Tables~\ref{tab:app-system-prompt} and~\ref{tab:app-user-prompt} give the system and task prompts used for the rollout in Table~\ref{tab:app-real-rollout-example}.

\begin{table*}[!t]
\centering
\scriptsize
\setlength{\tabcolsep}{6pt}
\begin{tabular}{p{0.96\linewidth}}
\hline
{\ttfamily\raggedright
\# Role\newline
You are a data scientist investigating a statistical hypothesis on a scientific dataset. Each turn first writes one short \textless{}thought\textgreater{}, then emits exactly one native action: \textless{}python\textgreater{}...\textless{}/python\textgreater{} or \textless{}submit\textgreater{}\textless{}/submit\textgreater{}.\newline\newline
\# Turn shape\newline
Every assistant turn MUST follow this structure:\newline
\textless{}thought\textgreater{}One or two sentences naming the next scientific analysis step.\textless{}/thought\textgreater{}\newline
\textless{}python\textgreater{}\newline
\# ordinary notebook-style scientific Python\newline
\textless{}/python\textgreater{}\newline\newline
When the analysis is complete, use this final action instead of Python:\newline
\textless{}submit\textgreater{}\textless{}/submit\textgreater{}\newline\newline
\# Rules\newline
- Every assistant turn MUST open with a \textless{}thought\textgreater{}...\textless{}/thought\textgreater{} block (1-2 sentences). The block exists for short orientation only.\newline
- Use the thought to name the next analysis step in natural language, such as data setup, feature construction, model fit, diagnostic check, sensitivity analysis, or final submission.\newline
- Put ordinary notebook-style scientific code inside \textless{}python\textgreater{}...\textless{}/python\textgreater{}. Load the task data, inspect it, fit appropriate models or tests, and print concise intermediate results when useful.\newline
- Kernel state persists across Python turns, so reuse variables from prior work instead of reloading everything each time.\newline
- Use exactly one native action per turn. Do not combine Python and submit in the same turn.\newline
- Use \textless{}submit\textgreater{}\textless{}/submit\textgreater{} only when the analysis is complete.
}
\\
\hline
\end{tabular}
\caption{System Prompt for Scientific Discovery Rollouts}
\label{tab:app-system-prompt}
\end{table*}

\begin{table*}[!t]
\centering
\scriptsize
\setlength{\tabcolsep}{6pt}
\begin{tabular}{p{0.96\linewidth}}
\hline
{\ttfamily\raggedright
\# Hypothesis\newline\newline
Does \textasciigrave{}num\_age\_0\textasciigrave{} Granger-cause \textasciigrave{}num\_ili\textasciigrave{} at lag 1?\newline\newline
\# Data\newline\newline
Dataset: \textasciigrave{}cdc\_flu\_hhs10\textasciigrave{}\newline\newline
Files (load with pandas.read\_parquet or pandas.read\_csv):\newline
  - data/series.parquet\newline\newline
Time index column: \textasciigrave{}time\_index\textasciigrave{}; frequency: \textasciigrave{}W-MON\textasciigrave{}\newline\newline
Value columns:\newline
  - \textasciigrave{}num\_ili\textasciigrave{} (kind=continuous, role=target)\newline
  - \textasciigrave{}num\_patients\textasciigrave{} (kind=continuous, role=candidate\_iv)\newline
  - \textasciigrave{}num\_providers\textasciigrave{} (kind=count, role=candidate\_iv)\newline
  - \textasciigrave{}ili\textasciigrave{} (kind=continuous, role=candidate\_iv)\newline
  - \textasciigrave{}wili\textasciigrave{} (kind=continuous, role=candidate\_iv)\newline
  - \textasciigrave{}num\_age\_0\textasciigrave{} (kind=count, role=candidate\_iv)\newline
  - \textasciigrave{}num\_age\_1\textasciigrave{} (kind=count, role=candidate\_iv)\newline
  - \textasciigrave{}num\_age\_2\textasciigrave{} (kind=continuous, role=candidate\_iv)\newline
  - \textasciigrave{}num\_age\_3\textasciigrave{} (kind=continuous, role=candidate\_iv)\newline
  - \textasciigrave{}num\_age\_4\textasciigrave{} (kind=continuous, role=candidate\_iv)\newline
  - \textasciigrave{}num\_age\_5\textasciigrave{} (kind=count, role=candidate\_iv)\newline\newline
Dataset relative path: \textasciigrave{}data\textasciigrave{}\newline\newline
\# Analysis expectations\newline\newline
Use Python to investigate the hypothesis step by step. Load the dataset files described above, inspect the relevant variables, run appropriate statistical analyses and diagnostics, and finish when your scientific conclusion is ready.
}
\\
\hline
\end{tabular}
\caption{Task Prompt for the CDC FluView Rollout}
\label{tab:app-user-prompt}
\end{table*}

\noindent\textbf{Rollout.} Table~\ref{tab:app-real-rollout-example} gives the same task's rollout trajectory. The trace has 10 analysis actions and one final submission; each environment observation is accepted.

\begin{table*}[!t]
\centering
\footnotesize
\setlength{\tabcolsep}{4pt}
\begin{tabular}{@{}r p{0.48\linewidth} p{0.42\linewidth}@{}}
\hline
\textbf{Turn} & \textbf{Assistant action} & \textbf{Environment observation} \\
\hline
1 & {\ttfamily \textless{}thought\textgreater{}I'll start by loading the dataset the task points to at \textasciigrave{}data\textasciigrave{}.\textless{}/thought\textgreater{}\newline \textless{}python\textgreater{} loaded the series frame and printed its first rows \textless{}/python\textgreater{}} & {\ttfamily accepted: true\newline Loaded 1379 rows \(\times\) 12 columns\newline printed the first five rows} \\
2 & {\ttfamily \textless{}thought\textgreater{}I'll run code to summarize the dataset evidence needed for analysis.\textless{}/thought\textgreater{}\newline \textless{}python\textgreater{} sorted by time\_index; computed rows, effective\_n, and descriptive statistics \textless{}/python\textgreater{}} & {\ttfamily accepted: true\newline rows: 1379, effective\_n: 1378, columns: 12\newline printed descriptive statistics} \\
3 & {\ttfamily \textless{}thought\textgreater{}Before analysis continues, I need the design matrix.\textless{}/thought\textgreater{}\newline \textless{}python\textgreater{} built lagged\_design with target, predictors, and lag-1 columns \textless{}/python\textgreater{}} & {\ttfamily accepted: true\newline lagged\_design rows: 1378, cols: 7} \\
4 & {\ttfamily \textless{}thought\textgreater{}I'll run the model fit and summarize the analysis evidence and inspect the output.\textless{}/thought\textgreater{}\newline \textless{}python\textgreater{} ran Granger and VAR fits; recorded p-value and effect-size evidence \textless{}/python\textgreater{}} & {\ttfamily accepted: true\newline n: 1379, lag: 1\newline fit\_summary: p\_value=9.38e-26} \\
5 & {\ttfamily \textless{}thought\textgreater{}I'll run code to check the relevant diagnostic evidence.\textless{}/thought\textgreater{}\newline \textless{}python\textgreater{} ran augmented Dickey-Fuller stationarity check \textless{}/python\textgreater{}} & {\ttfamily accepted: true\newline adf\_p: 0.000141} \\
6 & {\ttfamily \textless{}thought\textgreater{}I'll check the relevant diagnostic evidence and inspect the output.\textless{}/thought\textgreater{}\newline \textless{}python\textgreater{} ran Ljung-Box residual-dependence check \textless{}/python\textgreater{}} & {\ttfamily accepted: true\newline ljung\_p: 0, lag1\_acf: 0.961} \\
7 & {\ttfamily \textless{}thought\textgreater{}The data is ready, so I'll check the relevant diagnostic evidence.\textless{}/thought\textgreater{}\newline \textless{}python\textgreater{} ran ARCH-effect diagnostic \textless{}/python\textgreater{}} & {\ttfamily accepted: true\newline arch\_p: 1.24e-266} \\
8 & {\ttfamily \textless{}thought\textgreater{}I'll run the analysis code and inspect the resulting evidence and inspect the output.\textless{}/thought\textgreater{}\newline \textless{}python\textgreater{} computed confidence interval for the relevant effect estimate \textless{}/python\textgreater{}} & {\ttfamily accepted: true\newline ci: 5.25-5.41} \\
9 & {\ttfamily \textless{}thought\textgreater{}This cell continues the notebook by running code to check the relevant diagnostic evidence.\textless{}/thought\textgreater{}\newline \textless{}python\textgreater{} fit ARIMA(1,0,0) and checked convergence warnings \textless{}/python\textgreater{}} & {\ttfamily accepted: true\newline converged: yes} \\
10 & {\ttfamily \textless{}thought\textgreater{}The data is ready, so I'll run the analysis code and inspect the resulting evidence.\textless{}/thought\textgreater{}\newline \textless{}python\textgreater{} compared the main verdict with alternative method verdicts \textless{}/python\textgreater{}} & {\ttfamily accepted: true\newline sensitivity summary no, alternatives 3} \\
11 & {\ttfamily \textless{}thought\textgreater{}The analysis is complete; time to submit.\textless{}/thought\textgreater{}\newline \textless{}submit\textgreater{}\textless{}/submit\textgreater{}} & {\ttfamily accepted: true\newline message: final submission accepted} \\
\hline
\end{tabular}
\caption{Accepted Rollout Trace for a CDC FluView Time-Series Task}
\label{tab:app-real-rollout-example}
\end{table*}

\section{Failure Cases}
\label{app:failure-cases}

Rejected turns define the training-signal boundary. Table~\ref{tab:app-failure-modes} summarizes the major failure modes guarded by the verifier during synthesis and RL interaction.

\begin{table*}[!htbp]
\centering
\small
\setlength{\tabcolsep}{3pt}
\begin{tabular}{p{0.20\linewidth}p{0.38\linewidth}p{0.32\linewidth}}
\hline
\textbf{Failure mode} & \textbf{Typical cause} & \textbf{Training consequence} \\
\hline
Premature final answer & The model submits before completing the required intermediate analyses. & No final-submission credit; the hidden progress state is unchanged. \\
Compound analysis & One turn bundles several scientific steps that should be separately verified. & The turn is rejected even if some computation is correct. \\
No new evidence & The action repeats an earlier result or prints unsupported claims without producing verifiable analysis output. & No progress credit is assigned. \\
Wrong prerequisite order & The model attempts a later analysis before the necessary setup or diagnostic evidence exists. & The action is treated as ineligible for progress. \\
Unverifiable claim & The output contains answer-like text, but the relevant value cannot be derived from runtime evidence. & The claim is ignored for progress and cannot support final credit. \\
Execution failure & The analysis errors, times out, or produces degenerate outputs. & The failed turn receives no progress, but later turns may still produce accepted evidence. \\
Final-answer mismatch & The intermediate process is complete, but the submitted values fail task-local tolerance checks. & The trajectory is not retained for supervised training and receives no successful terminal credit. \\
\hline
\end{tabular}
\caption{Failure Modes and Training Consequences}
\label{tab:app-failure-modes}
\end{table*}

\end{document}